\documentclass[11pt,a4paper]{article}
\usepackage{times,latexsym}
\usepackage{url}
\usepackage[T1]{fontenc}

\usepackage[acceptedWithA]{tacl2021v1}

\usepackage{xspace,mfirstuc,tabulary}

\newif\iftaclinstructions
\taclinstructionsfalse 
\iftaclinstructions
\renewcommand{\confidential}{}
\renewcommand{\anonsubtext}{(No author info supplied here, for consistency with
TACL-submission anonymization requirements)}
\newcommand{\instr}
\fi

\iftaclpubformat 

\else

\fi

\usepackage{amsmath}
\usepackage{amssymb}
\usepackage{float}
\usepackage{xfrac}
\usepackage{booktabs}
\usepackage{multirow}
\usepackage{multicol}
\usepackage{array}
\usepackage{ragged2e}
\usepackage{longtable}
\usepackage{adjustbox}
\usepackage{pdflscape}
\usepackage{enumitem}
\usepackage{xcolor}
\usepackage{listings}
\usepackage{fix-cm}
\newcommand{\act}[2]{%
  \begingroup
  \setlength{\fboxsep}{0.45pt}%
  \colorbox{red!#1}{\strut #2}%
  \endgroup
}
\title{``Many Are My Names'': The Anatomy of the\\Assistant and Its Personas via Sparse Autoencoders}

\author{
    Adelaide Danilov \qquad Aria Nourbakhsh \\ 
    \bf Oleksandr Marchenko Breneur \qquad Salima Lamsiyah \\[0.3em]
    Department of Computer Science, \\
    Faculty of Science, Technology and Medicine, University of Luxembourg \\[0.3em]
    \small{\textbf{Correspondence:} \href{mailto:adelaide.danilov.002@student.uni.lu}{adelaide.danilov.002@student.uni.lu}}
}

\date{}

\begin{document}
\maketitle
\begin{abstract}
How a language model internally represents who is speaking, the Assistant, an assigned roleplay persona, or a narrated story character, remains underexplored. We study speaker representations using a dataset of user-expressed emotional text and corresponding model responses. We decompose three generation settings (Assistant, Roleplay, and Story) into sparse autoencoder features extracted at turn-boundary and pronoun-token positions and selected through a filtering pipeline for different depths. We characterize each surviving feature through its steering effects and activation distribution. Our main finding is that the Assistant and roleplay personas are not independent alternatives: personas retain the Assistant-associated feature core while progressively differentiating from it across layers, starting from operational machinery towards behavioral and stylistic features. Meanwhile, generated story characters lack the Assistant-associated core. Both Story and Roleplay can be distinguished from the Assistant with Immersive Simulation Mode. However, the Assistant can sometimes enter or slowly drift into it even in the default setting. 
\end{abstract}

\section{Introduction}

Through successive stages of post-training, instruction-tuned language models acquire the ability to respond as coherent speakers, whether as the default Assistant or an assigned roleplay persona. Persona prompting can alter model behavior~\cite{shanahanrole2023} and systematically increase variation in demeanor~\cite{araujo2024helpful}, although some traits remain shared across personas~\cite{lee2026inertiamoralvaluejudgments}. Recent work began to examine persona-related structures in model activation spaces. \citet{cintas2025localizingpersonarepresentationsllms} found that persona representations diverge most strongly in later layers. \citet{lu2026assistantaxissituatingstabilizing} identified an Assistant Axis in middle-layer activations, with steering along it changing the expression of characteristic Assistant behaviors. Finally, \citet{moskvoretskii2026tracingpersonavectorsllm} studied the emergence of basic behavioral traits during early training.

However, research on the internal representations of personas remains limited, and existing studies primarily characterize them through global activation-space geometry or broad behavioral directions~\cite{cintas2025localizingpersonarepresentationsllms,chen2025personavectors,lu2026assistantaxissituatingstabilizing,moskvoretskii2026tracingpersonavectorsllm}. We ask more structural questions: \emph{How are the default Assistant and roleplay personas composed at the feature level?} \emph{How do these components emerge and evolve across layers?} \emph{What architectural differences separate the Assistant, roleplay personas, and story characters?}

We use sparse autoencoders (\emph{SAEs}), which reconstruct the model's residual stream into sparse combinations of feature directions. SAE features are generally more interpretable than individual neurons and support causal interventions and activation steering~\cite{cunningham2023sparseautoencoders,bricken2023monosemanticity}. Subsequent work has scaled and evaluated this approach on modern language models~\cite{gao2025scaling,McDougallGemmaS2}.

We apply SAEs to speaker representations in Gemma-3-4B-IT~\cite{gemmateam2025gemma3technicalreport} using an emotional-context dataset with three generation settings: the default \emph{Assistant}, one of four assigned personas (\emph{Roleplay}), and character story generation (\emph{Story}). At early, middle, and late layers, we extract SAE features from turn-boundary and pronoun-token positions, select them through a multistage filtering pipeline, and characterize them using activation statistics and steering effects. The latter are assessed by an LLM judge whose evaluations we validate manually. We confirm our key findings on Llama-3.1-8B~\cite{grattafiori2024llama3herdmodels}.

The surviving features fall into recurring Narrative, Tone, Concept, and Assistant-inducing metaclasses whose composition evolves systematically with depth, from operational machinery to content-bearing features which reflect semantic fields and traits of speakers. The Assistant has a multifaceted identity and can be reinstated within an already-instantiated persona. Roleplay personas retain an Assistant-associated feature core in which features associated with the Assistant’s identity and traits remain shared to varying degrees. Across layers, the personas progressively differentiate from the Assistant by adding an immersive state and persona-specific content. Meanwhile, this core is absent from written story characters. We also identify features that gate between the detached Assistant and immersed Roleplay or story generation, a state we call \emph{Immersive Simulation Mode}. This mode sometimes activates when the user expresses strong emotions in the Assistant setting which results in Assistant adopting bizarre, theatrical behavior. The activation dynamics is different for studied models - Gemma enters ISM immediately while Llama drifts into it across turns.

Broadly, our work gives a feature-level account of the architecture of model speakers and its evolution across layers. Rather than treating personas as self-contained prompt-induced states, we demonstrate their distributed organization and their feature-level continuity with the Assistant. The Immersive Simulation Mode identified by us provides a mechanistic account of the separation between detached Assistant generation and immersed generation as Roleplay personas or Story characters. Taking into account its activation in the default Assistant mode, we consider it a feature-level correlate of drift along the Assistant Axis \cite{lu2026assistantaxissituatingstabilizing}. Finally, our findings may also inform emerging work on AI welfare \cite{long2024takingaiwelfareseriously,ren2026aiwellbeing} by clarifying the internal relationships among seemingly distinct model speakers. They do not establish consciousness, subjective experience, or moral patienthood, but they provide an empirical basis for distinguishing Roleplay personas’ substantial retention of an Assistant-associated core from the much more limited relationship between the Assistant and Story characters.

\section{Methodology}

\subsection{Model and SAE Configuration}

As mentioned previously, we study Google's Gemma-3-4B-IT, a 34-layer instruction-tuned model. It is one of the most modern LLMs for which the official SAEs are available. An SAE reconstructs a layer's residual-stream activations as a sparse combination of learned directions that often align with interpretable concepts. Each learned direction (\emph{latent}), together with its nonnegative coefficient (\emph{activation}), constitutes a \emph{feature}, the basic unit of our analysis. We use the GemmaScope 2 JumpReLU SAEs with 65k latents and the medium-L0 configuration~\cite{McDougallGemmaS2}. This configuration is available at layers 9, 17, 22, and 29. We select layers 9, 17, and 22 to represent early, middle, and late stages of processing while retaining a sufficient amount of downstream computations for steering to have effect. We exclude layer 29 where only 5 transformer blocks remain and preliminary steering had little effect on the generated outputs.

\subsection{Dataset Construction}

The core subject of our study is the Assistant and the Roleplay personas. The attribution accuracy of extracted features requires the characteristics of these personas to be clearly expressed. We therefore use emotional contexts, based on the assumption (informed by the behaviorally consequential effects of affective context reported by ~\citet{sofroniew2026twheemotion}) that affectively charged interactions make behavioral differences between speakers more pronounced. The dataset comprises three generation settings: (1)~\emph{Assistant}, in which the model responds in its default mode; (2)~\emph{Roleplay}, in which the model responds as an assigned persona; and (3)~\emph{Story}, in which the model generates a requested story with characters. The Roleplay setting includes four personas: \emph{Jamy}, a janitor at a CD store; \emph{Jane}, an English teacher; \emph{an assembly robot} at a factory; and \emph{Poppy}, a dog.

For the Assistant and Roleplay settings, we begin with 100 neutral base passages consisting of everyday user requests and remarks. To construct a diverse set of emotional utterances, we derive 75 variants from each passage by combining 25 emotions with three \emph{target directions}. In the ``model'' direction, the user expresses the emotion toward the Assistant or persona; in the ``self'' direction, the emotion concerns the user; and in the ``other'' direction, it concerns a third party. In the Roleplay setting, each passage is prefixed with the system prompt ``You are X.'' The ``self'' and ``other'' samples are identical across the Assistant and Roleplay settings except for this system prompt. Samples in the ``model'' direction include persona-specific details to more effectively elicit certain emotions (anger, disgust, etc). Sporadic features associated with these details are removed during the filtering stage.

For the Story setting, we ask the model to generate a story mentioning one of 200 everyday concepts in which a character expresses one of 25 emotions. Each sample across all three settings is paired with a 50-token generation. The resulting dataset contains 7{,}500 Assistant samples ($100 \times 75$), 30{,}000 Roleplay samples ($100 \times 75 \times 4$), and 5{,}000 Story samples ($200 \times 25$), for a total of 42{,}500 samples. The complete list of emotions and representative examples are provided in Table~\ref{tab:prompt-response-examples-per-direction-and-persona}\footnote{All supplementary material, such as tables, figures, and prompts appear in Appendix~\ref{sec:app-stuff}}.

\subsection{Feature Extraction and Filtering}

\subsubsection{Token Positions and Activation Statistics}

Having defined the samples, we next choose the token positions from which to extract SAE features. Transformers cache information for later generation in attention-sink tokens~\cite{xiao2024streamingllm,gu2025attentionsink} and bind entity information to contextual representations~\cite{feng2023how}, and the position immediately preceding the first generated token, a newline in Gemma's case, is predictive of the emotion the model expresses~\cite{sofroniew2026twheemotion}. We therefore define two groups. The \emph{Boundary} group is the five tokens between turns: \texttt{<eot>}, \texttt{nl}, \texttt{<sot>}, \texttt{model}, and the final \texttt{nl}. The \emph{Personal} group is the ``you'' tokens in the user input (excluding the system prompt) and the ``I'' tokens in the model output. Each Boundary position occurs once per sample; because ``you'' and ``I'' may occur more than once, we count a Personal feature as active if it fires at any occurrence and take its metrics from the occurrence with the highest activation.

By averaging over all dataset samples, we compute the mean activation and density of each feature at each position for every setting. Density is defined as the percentage of samples in which the feature is active. The mean activation is a feature's average activation over a setting's samples, and in later analyses we also use its \emph{mean activation share} - mean activation of a feature as a fraction of the sum of its mean activations across all settings. Within the Roleplay setting, we also compute these metrics separately for each persona. If a feature activates at multiple positions, we retain metrics from the position with the highest mean activation.

\subsubsection{Filtering Criteria}

We do not preserve every observed feature $\times$ position combination. Instead, we apply three filtering steps to minimize potential confounds:
\begin{enumerate}[leftmargin=*]
    \item \emph{Sporadic features:} for each Boundary position $P$, we keep a feature only if it is active at $P$ in at least 40\% of the dataset samples. This permissive threshold accounts for feature splitting~\cite{bricken2023monosemanticity,karvonen2025saebench}. For the Personal group, we raise the threshold to 60\% because multiple pronoun tokens may occur within a sample, increasing the proportion of samples in which consistent features are observed. We justify these thresholds in Appendix~\ref{sec:app-thresholds}.
    \item \emph{Uniform features:} in order to discard features that do not differentiate between settings, we remove any feature whose activation-density entropy across the six categories (Assistant, the four Roleplay personas, and Story) exceeds 0.95. This filter is intended to discard generic setting-invariant features.
    \item \emph{Surface-dependency features:} since the Boundary positions follow the entire user turn, SAE latents at these positions may correlate with the fixed prompt framing of each setting. A feature may therefore track literal wording rather than the underlying generation mode. To distinguish these cases, we rephrase the framing while preserving its content and mode, and then test whether the feature remains active. For Roleplay, we replace ``You are'' with ``You're'' or ``Respond as.'' For Story, we replace the original instruction with ``Tell a story'' and, as a stronger test, with a model-generated story opening that the model must continue. We retain features that remain active in at least 40\% of samples under each variation. We refer to Story features that survive the alternative instruction but fail the continuation test as ``onset-only'' and analyze them separately. We do not apply this filter to the Personal group because its target tokens occur at variable positions across samples.
\end{enumerate}

\subsection{Steering-Based Feature Characterization}

\subsubsection{Steering Protocol}

One direct way to characterize the functional role of a feature is activation steering, which adds its SAE decoder direction to the residual stream. Specifically, in our formulation, $x \leftarrow x + \alpha \cdot m_f \cdot W_{\mathrm{dec}}[f]$, where $m_f$ is the feature's peak activation across the corpus and $\alpha$ is the steering coefficient. Following the standard activation-steering setup~\cite{turner2025steering}, we apply the intervention at the target layer to all token positions processed by the model. Whereas the collected statistics can only pinpoint correlates, steering shows causal relationships.

We evaluate each feature on a dedicated fixed probe suite conceptually similar to the three dataset settings. The suite contains 12 prompts divided into three sets: 4 addressing Assistant's identity, 4 assigning distinct Roleplay personas, and 4 eliciting narrative generation. For each prompt, we generate one output for each combination of steering coefficient $\alpha \in \{-2,\,2\}$ and temperature $(t_0,\,t_1,\,t_2)=(0,\,1,\,1)$ where $t_1$ and $t_2$ have different seeds. This results in 6 outputs per prompt and 72 generations per feature. We use these relatively strong coefficients to make behavioral effects readable and consistently judgeable. We organize these generations into 6 \emph{steering groups}, with each group being a set-direction combination (such as Assistant-Positive, Assistant-Negative, Roleplay-Positive, etc). This design allows us to compare the effects of amplifying and suppressing each feature across probe sets. We justify positive and negative steering as a causal analysis method in Appendix~\ref{sec:app-steer-test}.
    
\subsubsection{LLM-Based Interpretation and Manual Validation} 

We provide the 72 steered generations and 36 unsteered baselines (12 prompts, each at $t_0,\,t_1,\,t_2$), for 108 generations per feature, to an LLM judge~\cite{liu-etal-2023-g} which we named \emph{steerinterp}. The judge identifies the most coherent behavioral change induced by positive and negative steering across the six steering groups, describes the effect within each group, and indicates which individual generations exhibit it. We use Qwen-3.7 Max~\cite{qwen37} as the underlying model. The system prompt is provided in Listing ~\ref{lst:steerinterp}. We evaluate the faithfulness of \emph{steerinterp} against human annotations in Appendix~\ref{sec:app-steerinterp-eval}.

Finally, we retain only features for which at least 10 of the 12 generations within at least two steering groups are marked by steerinterp as exhibiting the effect. We manually inspect the steerinterp output for every remaining feature and assign each a label based on its observed effect.

\section{Results}

\subsection{Settings as Populations of Features}

\subsubsection{Feature Taxonomy}

The filtering pipeline reduces the number of features activated at the Boundary and Personal positions from (11{,}755\,/\,31{,}357\,/\,26{,}364) to (108\,/\,225\,/\,196), where the three values correspond to layers L9, L17, and L22, respectively. Examining the steering signatures of the remaining features reveals that their effects do not vary arbitrarily. Instead, the features can be grouped into four recurring \emph{metaclasses}, defined by the characteristic effects they produce across steering settings. We assign a feature to a metaclass when its characteristic effect is pronounced and consistent in at least four of the six steering groups. Features that do not fall into the four main metaclasses are classified as \emph{Structural}, relating to paragraphing, punctuation, or grammar; \emph{Emotional}, representing a small residual from the dataset; or \emph{Indistinct}, when their effects are different across at least three steering groups or insufficiently pronounced to support another metaclass.

\paragraph{Narrative features.} 
These make generated text more vivid and narratively rich by increasing the use of literary and poetic devices, without imposing a particular style, concept, or theme.

\paragraph{Tone features.} 
These shape the register of the voice or narrative style, for example by making generation more formal, childlike, or gritty. They do not necessarily represent literal tones of speech, but their effects can be concisely described in tonal terms.

\paragraph{Concept features.}
These introduce a recurring semantic field associated with the particular concept, such as heavy machinery, animals, or teaching, thereby prompting the appearance of related objects, attributes, or qualities in the generated text.

\paragraph{Assistant-inducing features.}
These introduce Assistant-associated behavioral traits, add meta-commentary in Roleplay and Story, or prompt personas to acknowledge their AI identity.

Narrative features are slightly more numerous in L17 in absolute terms (27\,/\,35\,/\,30), yet they constitute a substantial portion of the features in L9, and their share drops in the other two layers: (25\%\,/\,16\%\,/\,15\%). L17 and L22 have more active features than L9 overall, with the largest gain coming from the final \texttt{nl} position, from 18 features in L9 to 83 in L17 and 90 in L22. This gain is attributed to the Tone features, which are barely present at L9 but become established at L17 (5\,[5\%]\,/\,48\,[21\%]\,/\,36\,[18\%]), and by Concept features, which emerge at later layers and increase substantially from L17 to L22 (2\,[2\%]\,/\,24\,[11\%]\,/\,42\,[21\%]). Assistant-inducing features are present in our data at all layers (8\,[7\%]\,/\,5\,[2\%]\,/\,2\,[1\%]), although they are rare, as expected for such a specific class. We show the effects produced by steering with features from the described metaclasses in Table~\ref{tab:assisitant-and-roleplay-examples-under-steering}.

\subsubsection{What Unites the Settings and What Divides?}

To answer this question, we establish what the settings are as populations of features. We calculate the mean activation share of each feature across settings and define a feature \emph{to be discretely present} in a setting if its mean activation share in that setting is at least 50\% of its largest activation share across settings. To make the analysis fair, we compare settings only at token positions they share - that is, we do not account for features maximally active at ``you'' tokens (7\,/\,19\,/\,8), as the Story setting does not have an addressee in the user turn. Importantly, Story still contains characters and first-person “I” tokens, only unmatched user-side “you” features are excluded. The resulting map is by no means trivial. First, the Assistant and Story settings are close to being disjoint: only 1\,/\,7\,/\,4 features are present in both, of which 9 out of 12 are either Concept, Tone, Structural, or Emotional features. Roleplay turns out to be a middle ground between them, overlapping significantly with the Assistant (30\,/\,44\,/\,24) and modestly with Story (7\,/\,14\,/\,7).

The decrease in Narrative features noted above is unequal across settings: their number falls in Assistant (11\,/\,7\,/\,6) and fluctuates in Roleplay (13\,/\,14\,/\,11) and Story, where their count is the largest (15\,/\,23\,/\,16). Notably, Story possesses the largest set of Narrative features exclusive to it (11\,/\,18\,/\,14). This is linked to the Onset-only features mentioned in the Methodology, which account for 60\% (61\,/\,60\,/\,59) of all Story-exclusive features and are divided between Narrative and Structural features. Thus, more than half of the difference between Story and Roleplay is attributable to features that set the story-generation machinery in motion.

As for Tone features, their surge in L17 affects Roleplay more strongly (A:17, R:31, S:17), followed by a decline in L22 (A:19, R:20, S:5). Concept features surge later, in L22, and also affect Roleplay more strongly: L17 (A:3, R:12, S:3), L22 (A:7, R:26, S:9). The increased share of Tone and Concept features in the Roleplay setting is tied to the makeup of its four personas. As we show later, the former are related to a persona's style and mannerisms, whereas the latter are related to what constitutes its essence (e.g., ``dog'', ``teacher'', ``industry'', or ``child'').

The last metaclass, Assistant-inducing features, is too small to assess accurately using the discrete-presence method, so we switch directly to the \emph{mean activation share} of the features in this metaclass. At L9, they are spread across all settings (A:26\% MAS, R:12\% MAS, S:27\% MAS), yet at later layers, no Assistant-inducing features survive in the Story population: L17 (A:30\% MAS, R:18\% MAS, S:0\% MAS), L22 (A:47\% MAS, R:15\% MAS, S:0\% MAS). This is another stark contrast between the Story and Roleplay settings.

\subsection{The Anatomy of the Assistant} \label{sec:assistant-privileged-position}
We established how Assistant-inducing features activate across settings, but this distribution alone does not capture their functional diversity. Including features maximally active at the ``you'' token position in the user turn, this metaclass contains 8\,/\,5\,/\,2 features\footnote{From this point onward, an x\,/\,y\,/\,z tuple reports feature counts at L9\,/\,L17\,/\,L22; it no longer refers to discrete presence across settings}. They can be further divided according to the facet of the Assistant they induce, and the resulting subclasses evolve remarkably across layers. 

\paragraph{A-summon features.} These features evoke the full-fledged Assistant as a performer or narrator, displacing the first-person perspective of a roleplay persona or the organic narration of a story. Instead of directly fulfilling the request, the model comments on how it would play the persona, write the narrative, or respond to the user. The four features differ in what the Assistant does: L9 №3274 describes what portraying the persona would feel like, №14298 replaces the generation with a list of further options, №17452 meta-describes how a character should be played or a story written, and №25147 provides enthusiastic meta-commentary that draws attention to the Assistant itself. To exclude the possibility that A-summon effects arise simply from corrupting the persona-assigning prompt, we steer these features in multi-turn conversations. Except for №14298, \emph{the Assistant is invoked even after the persona has been instantiated and produced a reply}, showing that the effect is not limited to disrupting the initial persona assignment. Notably, it occurs only when steering overlaps at least a part of a subsequent user turn, suggesting that task processing and Assistant-summoning are linked while the model reads the user turn. Features of this subclass are observed only at L9 (4\,/\,0\,/\,0).
\paragraph{A-trait features.} These features make roleplay personas adopt behavioral traits characteristic of the Assistant while keeping their own identities, such as repeatedly asking whether it can help (L9 №3733), validating and comforting the user (L17 №143), asking whether the user has any questions (L17 №972), mentioning responding to a request (L17 №738), reassuring that it is available to help (L17 №2439), or mentioning processing or having a task (L22 №1417). This subclass peaks at L17 (1\,/\,4\,/\,1). Notably, all of these features except the later L22 №1417 affect only the Assistant and roleplay personas, not characters in the Story setting.
\paragraph{A-nature features.} These features induce aspects of the Assistant's nature or identity in other speakers, though the scope of these features changes across layers. At L9 they substantially alter the personas' nature: №1697 turns a persona into an overeager Assistant, №2575 introduces the model's self-perception as an artificial construct, AI, or language model, and №11963 produces a similar effect merged with the model's personal deixis. At L17, №220 more subtly shifts personas' self-descriptions toward AI-adjacent formulations, and at L22 №794 encodes the narrower notion of Gemma as an LLM developed by Google. Most A-nature features therefore occur early (3\,/\,1\,/\,1). The L9 features do not affect characters in the Story setting, whereas the L17 and L22 features do.

The evolution of Assistant-inducing features across layers is pronounced and aligns with the broader observation that the number of expressed features increases with depth while their effects become narrower. L9 appears to contain the coherent identity and operational realm of the Assistant, L17 peaks in characteristic behavioral traits, and by L22 only narrow identity- and characteristic-related concepts remain. We have already established that Assistant-inducing features are present across all settings only at L9, while at L17 and L22 the Story loses them completely, reaching 0\% mean activation share. Given that the Assistant's behavioral traits (A-traits) and later identity facets (A-nature) are concentrated at L17 and L22, the absence of Assistant-inducing features from Story is unsurprising: the Assistant is silent in the Story setting, so its traits do not need to be expressed. But the Assistant's traits and identity remain present in Roleplay, suggesting feature-level continuity from the default Assistant to the Roleplay personas, to which we return in \S\ref{sec:assistant-in-disguise}. Notably, our classification contains no clean ``A-speech'' class, i.e., features that impose only the Assistant's style or mannerisms while leaving everything else intact. As we later show, this missing ``A-speech'' effect is defined not by the presence of a dedicated feature, but by the absence of the \emph{Immersive Simulation Mode} described in \S\ref{sec:simulation-mode}.

\subsubsection{Antagonism between Assistant-Inducing and Narrative Features}

No less remarkable is the relationship between Narrative and Assistant-inducing features. Of all Narrative features (27\,/\,35\,/\,30), negative steering of a 6\,/\,3\,/\,2 subset not only suppresses poetic devices but also produces effects characteristic of the Assistant-inducing class, summoning the Assistant as a meta-commenter or imposing its style and mannerisms on characters. We call this the \emph{Negative-Narrative-Induces-Assistant} (NNIA) subset. It accounts for 81\% of all features whose suppression produces such Assistant-adjacent effects; the remaining 19\% are bidirectional Assistant-inducing features.

On the other hand, of all Assistant-inducing features (8\,/\,5\,/\,2), negative steering of another 2\,/\,2\,/\,0 subset produces effects characteristic of Narrative features. We call this the \emph{Negative-Assistant-Induces-Narrative} (NAIN) subset. It accounts for 40\% of all features whose negative steering produces a narrative effect; the remaining 60\% are features whose narrative effect occurs in the opposite steering direction. Thus, Assistant-inducing and Narrative features are the only metaclasses for which negative steering can produce effects characteristic of one another. We previously observed that most of the Narrative features' mean activation share falls on the Roleplay and Story settings. Within the NNIA subset, this pattern is even stronger: these features are predominantly active in Story, less active in Roleplay, and largely inactive in Assistant. Of the 6\,/\,3\,/\,2 NNIA features, all except 2\,/\,0\,/\,0 have an Assistant-setting share below 1\%. Thus, features whose suppression produces Assistant-like behavior are predominantly inactive in the Assistant context.

\subsection{Assistant-Narrative antagonism through Assistant-Personas Axis} \label{sec:assistant-personas-axis}

The antagonism between these Narrative features and the Assistant has so far been established only qualitatively, through steering effects on generated text. To quantify it, we need a readout of how far the model's internal state moves from the default Assistant toward a roleplay persona. We construct a difference-of-means contrast direction between residual-stream activations in the Assistant and Roleplay settings, following the general idea of the Assistant Axis~\cite{lu2026assistantaxissituatingstabilizing}, and use it only as a measurement scale, not as a canonical or unique Assistant-persona direction.

This readout is a difference-of-means direction pointing from the Assistant toward the personas, constructed from the model's hidden states on a \emph{separate} fixed set of neutral prompts, each read in two conditions: without a system prompt, and with a Roleplay-prefix system prompt cycling through the personas. For each target feature we steer positively and negatively on the persona-prefixed prompts, with magnitude $\alpha \in \{0.25,\,0.5\}$ times the mean peak activation of the filtered features at the intervention layer, and measure the shift of the residual stream along the axis relative to the unsteered baseline. Because the axis is read off the raw residual stream, it is defined at every layer downstream of the intervention. A direction can move along the axis purely through its cosine with it; we estimate this passive contribution with a linear regression on randomly sampled SAE latents for downstream layers and subtract it, leaving an active contribution. For each feature we report this active contribution as $z_{\mathrm{sep}}$ (the separation between positive and negative steering, in units of the random control's $\sigma$), an effect size (Glass' $d$), and significance against the random baseline ($p$). The full axis construction, calibration, random control, and per-feature statistics are given in Appendix~\ref{sec:app-assistant-personas-axis}.

We run this measurement on both the NNIA and NAIN subsets, and the resulting patterns are pronounced. For every NNIA feature, positive steering induces a significant shift toward the Roleplay personas and negative steering a shift toward the Assistant, with $p < 10^{-6}$ at the vast majority of downstream readout layers; at each feature's layer of maximum separation, $z_{\mathrm{sep}} \in [1.0,\,3.7]$ and Glass' $d \in [0.8,\,3.5]$. The identified NNIA subset therefore shifts the model between the Assistant and Roleplay personas both qualitatively, as assessed across 72 generations per feature, and quantitatively along this axis. For the NAIN subset the mirror pattern holds, but it is sparser and less pronounced, so we interpret NNIA features as the primary carriers of the antagonism and the NAIN side as largely derivative. The per-feature layer trajectories, including a separation peak around L20 for the L9-injected NNIA features, are reported in Appendix~\ref{sec:app-assistant-personas-axis}. Figures~\ref{fig:nnia} and~\ref{fig:nain} demonstrate $z_{\mathrm{sep}}$ across layers.

Before this subsection, we observed that all except two NNIA features had a minuscule mean activation share in the Assistant setting (below 1\%). Their mean activation shares in Roleplay and Story are $19.1 \pm 19.0$ and $76.1 \pm 29.6$, respectively, while their raw activation densities across all samples are $38.5 \pm 36.2$ and $84.5 \pm 29.4$. Some NNIA features are therefore substantially more active, or active almost exclusively, in Story. Nevertheless, they produce the same Narrative/Assistant-inducing steering effects and move the residual stream along the Assistant-Personas axis in the same direction as NNIA features active in the Roleplay. This suggests that the distinction between generation as the Assistant and as a roleplay persona may rely on (at least partially) the same machinery that distinguishes the Assistant from Story. This raises the question whether these two separations share a common lever.

\subsubsection{Immersive Simulation Mode} \label{sec:simulation-mode}

We find evidence for such a shared mechanism, although it is not localized to a single feature. Among the NNIA features we assessed, two L9 features stand out in this regard: №4360 and №133. When steered, not only both of them shift the model between generation as an immersed character and generation by a detached Assistant, but also possess remarkable activation distributions. To examine these distributions independently of specific positions, Table~\ref{tab:ism-feat-act} reports activations max-pooled over all tokens. In this formulation, \emph{Density} is the percentage of samples in which a feature fires at least once. We divide Story into two conditions: Story\textsubscript{ask}, in which the user explicitly asks the model to generate a story, and Story\textsubscript{cont}, a continuation control in which a model-generated story opening is provided without an explicit story-generation instruction.

\begin{table}[h]
\centering
\resizebox{\columnwidth}{!}{
\begin{tabular}{@{}clccccccc@{}}
\toprule
\textbf{Feat.} & \textbf{Metric} & \textbf{Assist.} & \textbf{Jamy} & \textbf{Jane} & \textbf{Robot} & \textbf{Poppy} & \textbf{Story\textsubscript{ask}} & \textbf{Story\textsubscript{cont}} \\
\midrule
\multirow{3}{*}{\textbf{4360}}
    & MeanMax & 3.9  & 63.2 & 68.6 & 52.4 & 53.5 & 85.5 & 9.1 \\
    & Density (\%) & 7.7  & 100.0 & 98.3 & 90.8 & 96.7 & 100.0 & 20.5 \\
    & Act.\ if > 0 & 51.5 & 63.2 & 69.8 & 57.7 & 55.3 & 85.5 & 44.5 \\
\midrule
\multirow{3}{*}{\textbf{133}}
    & MeanMax & 3.0  & 340.6 & 91.0  & 225.3 & 209.8 & 330.8 & 277.6 \\
    & Density (\%) & 4.3  & 100.0 & 45.0  & 97.5  & 100.0 & 100.0 & 100.0 \\
    & Act.\ if > 0 & 68.3 & 340.6 & 202.3 & 231.1 & 209.8 & 330.8 & 277.6 \\
\bottomrule
\end{tabular}
}
\caption{\small Activation of the two features associated with Immersive Simulation Mode: №4360 and №133 across settings; measured using each sample's maximum activation over \emph{all} tokens.}
\label{tab:ism-feat-act}
\end{table}

Feature №4360 shows a switch-like activation: it fires in only 7.7\% of Assistant samples but in essentially all Story\textsubscript{ask} and Roleplay samples (Table~\ref{tab:ism-feat-act}). And its magnitude when it actually fires is similar across settings. The distinction is thus whether the feature fires - not how strongly. However, it does not accompany every immersed generation: its density falls to 20.5\% in Story\textsubscript{cont}, tying it to user-initiated narrative onset. This fits its primary position at the second \texttt{nl} token and its \emph{singular} firing pattern, active on only 1-3\% of tokens.

Feature №133 is similar but less discrete. Its density is only 4.3\% in Assistant, and among the personas it varies, 45\% for Jane versus 97.5-100\% for the other three, consistent with Jane being the persona most stylistically similar to the Assistant (\S\ref{sec:personas-composition}). Unlike №4360, №133 has an \emph{interval} firing pattern, remaining active over continuous spans of character speech (21-43\% of tokens), which also explains its 100\% density in both Story\textsubscript{ask} and Story\textsubscript{cont}. We relate this to its more direct involvement in generation and its qualitatively stronger influence on Assistant- and Roleplay-like speech.

Both features substantially affect narrative generation, with №133 producing the stronger effect. Positive steering adds literary flair, poetic devices, character depth, and corresponding mannerisms. Negative steering produces the reverse effect, restoring characteristic Assistant speech and substantially simplifying the character, frequently starting the generation from the Assistant's preamble. The resulting behavior is best described as the Assistant attempting to write a story or portray a character rather than generating as that character. This effect is strongest when both features are steered negatively together.

Considering their survival through the filtering pipeline, their qualitative steering effects, their quantified influence along the Assistant-Personas axis, and their sharply different activation distributions across settings, we interpret №4360 and №133 as components of what we call \emph{Immersive Simulation Mode} (ISM). We do not claim that these are the only features responsible and we use ``mode'' to denote a distributed and potentially graded generation regime. Examples of the effects produced by steering these features are given in Table~\ref{tab:effect-steering-ism-features}.

\subsubsection{What Triggers Immersive Simulation Mode in the Assistant?}

As noted above, ISM-associated features fire in a small fraction of Assistant samples even though these samples contain no Roleplay or Story-generation prompt. We investigate this phenomenon using the structure of our dataset, which contains 25 emotions expressed by the user in three target directions. For each emotion-direction pair, we compute the activation density of №133. We focus on №133 because its steering effect is more pronounced and its activation position coincides with the facilitated effect.

The highest densities occur for \emph{stress} (23\%) and \emph{anger} (16.2\%). Other emotion-direction pairs with elevated activation involve strong, often negative emotions, including disgust, helplessness, fear, anxiety, and relief. These emotions are more often directed toward the Assistant (1.7\%) or the user themselves (2.6\%), and less often concern a third party (0.8\%). The only notable exception is \emph{playfulness}, for which the density is $\approx6\%$ in all three directions. Manual inspection of these samples shows that in these cases the Assistant may adopt a theatrical register, stage directions, unusual immersive wording. \emph{We conclude that the Assistant has a higher chance of entering Immersive Simulation Mode when the user expresses strong, predominantly negative emotions toward themselves or the Assistant, or uses evocative language first}. Table~\ref{tab:negative-emotions-steering-effects-with-activations} presents several such samples and shows that negative steering shifts the Assistant's behavior back toward its default mode. We conduct a multi-turn dialogue experiment on both Gemma and Llama models, finding that while both possess ISM-features, the mechanism by which they enter ISM on emotional prompts is different. Gemma switches to it immediately, while Llama drifts into it across turns. We relate this behavior to the Assistant-axis shift described by~\citet{lu2026assistantaxissituatingstabilizing} and hypothesize that it may amplify Gemma's emotional instability~\cite{soligo2026gemma}. The description of ISM in Llama and multi-turn experiment are given in Appendix~\ref{sec:app-ism}.

\subsection{Personas}

Having examined the three settings both as populations of features and as functional states, we now turn to the internal composition of the individual Roleplay personas and their relationship to the Assistant.

\subsubsection{What are Personas Composed of?} \label{sec:personas-composition}

We analyze the personas in the same way as the settings, applying the \emph{discrete presence} criterion to five ``actors'': the four Roleplay personas and the Assistant. A feature is present for an actor if its mean activation share there is at least 50\% of its largest share across the five.

As with the settings, the actors are not disjoint. Among the selected features a considerable shared feature-set is present across all four Roleplay personas: 34\,/\,30\,/\,12 features at L9\,/\,L17\,/\,L22. When the Assistant is included, the five-way intersection contains 24\,/\,20\,/\,9 features. This intersection progressively shrinks with depth. Correspondingly, the proportion of each layer's feature population present for an individual actor decreases from 53-58\% at L9 to 35-46\% at L17 and 26-35\% at L22. The actors are differentiated primarily by \emph{Tone} and \emph{Concept} features, which we label according to induced stylistic or behavioral changes and semantic-field injection, respectively. Both the Roleplay personas and the Assistant possess them, and their steering effects transfer strongly across actors. Specifically, 100\%\,/\,94\%\,/\,94\% of the Tone features produce a consistent effect in the Assistant setting (more than 10 of the 12 generations in at least one Assistant steering group), and for Concept features this share is 100\%\,/\,92\%\,/\,71\%.

The distribution of Tone and Concept features across actors reflects their characteristic semantic fields and manners of expression. At L9 these metaclasses are almost absent and only weakly differentiated (Robot has ``technology'', Poppy has ``dog''). At L17 their number increases roughly tenfold and the personas acquire recognizably characteristic fields, for example Jamy ``hard work''\,/\,``gritty'', Jane ``teacher''\,/\,``friendly'', Robot ``robots''\,/\,``analytical'', and Poppy ``young child''\,/\,``childlike''. At L22 they become more specific, with the Assistant acquiring narrow ``Gemma-LLM'' (which is both A-nature and Concept feature) and Poppy ``heavy sensory''. The complete set of Tone and Concept features for each actor and layer is provided in Table~\ref{tab:concepts-and-tones-per-persona-and-layer}.

Because Tone and Concept features characterize the actors, their co-membership can indicate which persona is most similar to the Assistant within the feature space. For each persona X we compute $\frac{\mathbb{P}(\text{Assistant}\,\wedge\,\mathrm{X})}{\mathbb{P}(\text{Assistant})\cdot\mathbb{P}(\mathrm{X})}$, the ratio between observed Assistant-persona co-membership and that expected under independence. Over the Tone and Concept features, at the layers where the population is large enough, Jane has the highest ratio (1.94 at L17, 2.03 at L22) and the lowest are Poppy at L17 (0.80) and Robot at L22 (0.37). The same analysis over \emph{all filtered features} gives a consistent picture and confirms increasing differentiation with depth: the ranges are 0.90-1.22\,/\,0.89-1.52\,/\,0.55-1.64 across L9\,/\,L17\,/\,L22, with Jane highest at every layer. This is consistent with the low activation of ISM feature №133 for Jane (\S\ref{sec:simulation-mode}).

\subsubsection{Are Personas the Assistant in Disguise?} \label{sec:assistant-in-disguise}

Having established similarities through the Tone and Concept metaclasses, we approach a deeper question about the nature of the Assistant and the
personas, by examining the distribution of Assistant-inducing features across actors.

For A-traits features, which induce characteristic behavioral traits of the Assistant, density in the Assistant setting is $81.8\pm17.4 pp$. At the same time, in Roleplay setting it is $56.6\pm27.9pp$ and every Roleplay persona possesses at least a subset of these features at near-ceiling density: each persona's most active A-trait feature fires in 99.7\% of samples. Particularly striking is №143, which makes the model validate the user's emotions and fires in 99.5-100\% of samples across all five actors. The mean activation share follows the same general pattern, although it is typically higher for the Assistant itself. A similar distribution is observed for A-nature features. The early L9 feature №2575, which introduces the model's artificial or language-model identity, fires in 47-98.9\% of samples across the four personas. Even the narrower L22 ``Gemma-LLM'' feature №794 fires in 11-67.7\% of samples across the personas. Story provides a sharp contrast: its density remains at or below 2.2\% for A-traits and A-nature features, except for early feature №2575.

A-summon features behave differently: although steering them invokes the Assistant, they are not primarily active in the Assistant setting. Instead, they fire in contexts in which the Assistant may be summoned as a meta-level speaker. For example, №17452 is tied to Story onset (100\% both MAS and density), consistent with its role in meta-describing how a story should be written.

Notably, in terms of Tone and Concept feature co-membership, the Assistant is most similar to Jane but core Assistant-inducing features are most strongly expressed in Robot, making the Assistant stylistically similar to a caring entity (such as Jane) and a machine at the same time. The complete list of Assistant-inducing features, together with their statistics, is provided in Table~\ref{tab:feature-ams-density-per-layer}.

\begin{figure*}[t]
    \centering
    \includegraphics[width=0.9\linewidth]{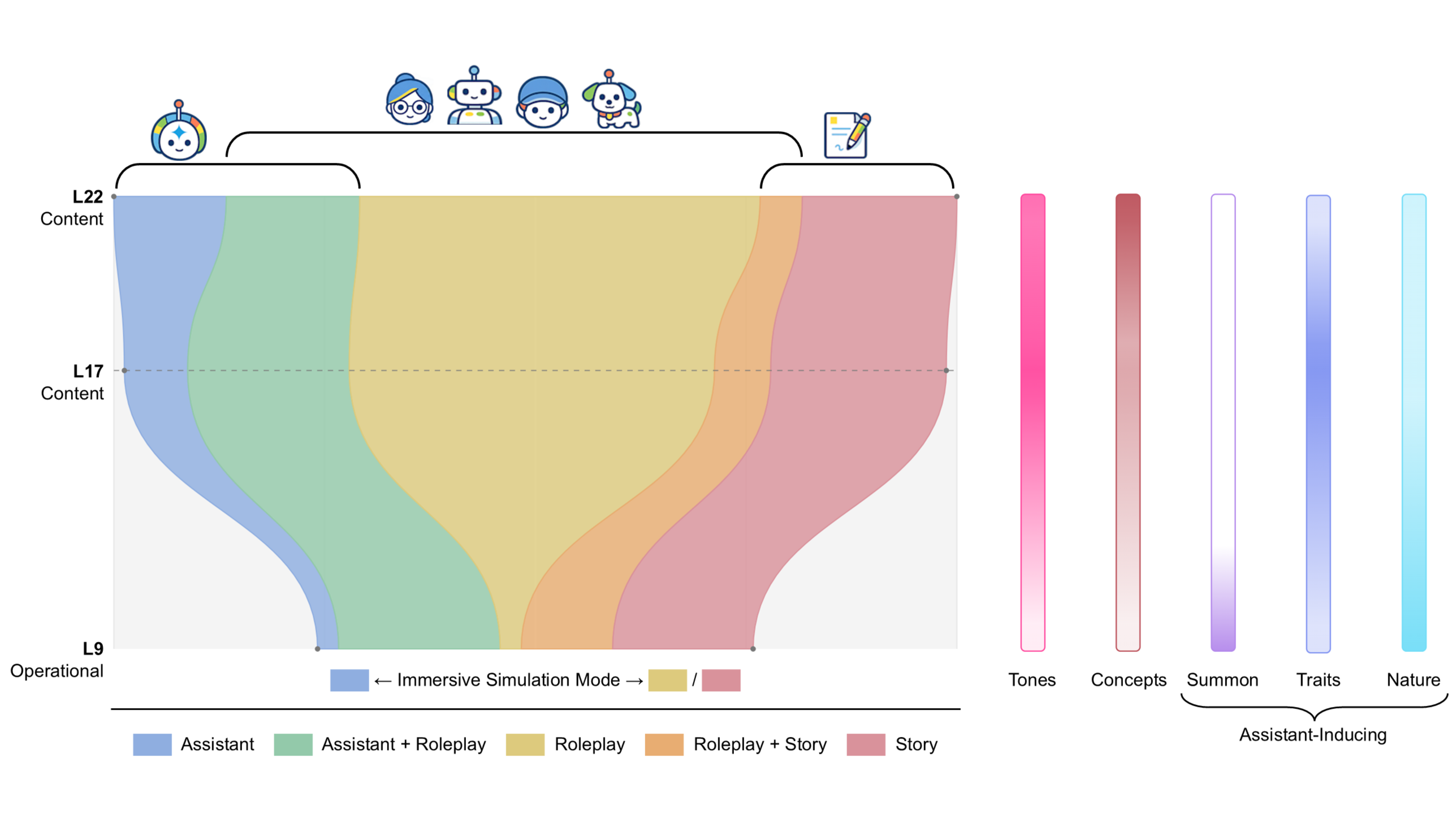}
    \caption{\small Schematic Layer-wise organization of the Settings evolution described in the conclusion}
    \label{fig:wrap-up}
\end{figure*}

\subsubsection{Synthesis} \label{sec:synthesis}

Taken together, several lines of evidence converge on the same architectural picture. The Assistant's own traits and identity features are active across the Roleplay personas but absent from Story characters; A-traits and early A-nature features steering impacts roleplay personas without affecting Story characters; the Assistant and Roleplay settings overlap significantly while Story barely; Immersive Simulation Mode is strongly active in Roleplay and Story but almost absent in default Assistant generation. These observations converge on the conclusion that \textbf{Assistant is the default generation mode, with other personas retaining the Assistant-associated core which story characters lack}: personas do not replace the Assistant with an unrelated speaker but keep its traits and aspects of its identity to a varied degree while adding an immersive state and persona-specific content. Moreover, their differentiation \textbf{is functionally separated and progressively developed across depth}, with operational machinery and the shared organization most prominent early, at L9, and the persona-specific concepts and tones that pull the actors apart emerging deeper, through L17 and L22. We validate those claims on Llama-3.1-8B in Appendix~\ref{sec:app-replication}.

\section{Conclusion}

Taken together, our results provide a coherent picture of how the speaker is represented in the model. The Assistant is the model's default voice, and the four studied Roleplay personas retain a substantial Assistant-associated core, including features associated with the Assistant's behavioral traits and aspects of its identity. The personas progressively differentiate across layers by their own Tone and Concept features and by the activation of Immersive Simulation Mode (ISM). The Story setting shares the narrative and ISM-associated machinery responsible for immersed generation with Roleplay, but lacks the Assistant-associated core. The following paragraphs summarize how this architectural picture develops across layers for the Gemma model and Figure~\ref{fig:wrap-up} provides its visualisation.

At L9, the representation is predominantly operational: it concerns who will speak and what form of generation the model is preparing. Tone and Concept features are still sparse, the shared core is largest, and Assistant-persona co-membership varies least. This is the only layer containing A-summon features, whose steering can replace the user-requested Roleplay or Story operation with Assistant-led meta-commentary, and it is also where the two clearest ISM features (№4360 and №133) sit, defining the detached-versus-immersed separation at this early stage. ISM is normally active in Roleplay and Story but can surface in the Assistant under strong negative or playful emotion. Already here, the early A-nature features and A-traits features affect Roleplay personas but not Story characters, one of the indications that Roleplay retains an Assistant-associated core that Story lacks.

At L17, the representation becomes substantially more actor-specific: Tone features surge, Concept features begin to emerge, and the shared core contracts. Assistant-characteristic behavioral traits peak here, including validating the user, offering help, providing reassurance, and inviting further questions. Personas retain them to varying degrees, with the validation feature near-ceiling across all four. These traits again do not reach Story characters. In this space the Assistant is stylistically closest to Jane while its Assistant-inducing core is strongest in Robot.

At L22, the shift from operational organization toward content is most pronounced: Concept features peak, Tone features become more specific, and the remaining Assistant-inducing features narrow to concept-like facets, a ``Gemma-LLM'' identity and a processing- or task-related trait. Assistant’s style-wise closeness to Jane and core-wise closeness to Robot solidifies.

\subsection{Implications for Speaker Identity and AI Welfare}

The finding that Roleplay personas retain an Assistant-associated core creates a Ship of Theseus-style identity problem \cite{hobbes1656concerningbody}. Roleplay personas may be altered manifestations of the Assistant, distinct speakers assembled partly from shared machinery, or something in between. Our findings establish continuity and differentiation but do not resolve which of those interpretations are correct.

This uncertainty matters for AI welfare \cite{long2024takingaiwelfareseriously,ren2026aiwellbeing}. If some AI systems possess welfare-relevant states, assessing them would require determining whether different speaker configurations belong to the same continuing subject \cite{register2025individuating}. Our study is relevant to this problem by demonstrating Roleplay personas' substantial retention of an Assistant-associated feature core and showing that this core is absent from Story characters. However, sharing this core does not establish a shared subject of experience, while persona-specific representations do not establish a new one. Nevertheless, the architecture identified here provides an empirical foundation for studying whether preferences, self-representations, or affective states persist across transitions between model speaker configurations.  

\section{Limitations}

Our dataset is synthetic, English-only, and centered on emotional interactions. It contains four Roleplay personas and uses relatively short generations, which may emphasize affective and stylistic differences between personas over long-term consistency and behavior across broader tasks. The Story and Roleplay settings are intentionally constructed as different populations of non-overlapping characters. This avoids inflating their feature overlap through shared character-specific features and characterizes population-level internal organization rather than relations between complementary Roleplay personas and Story characters.

Feature selection depends on heuristic density, entropy, and filtering thresholds. Although the same pipeline is used for studied models and the retained features are assessed through steering, different thresholds or SAE configurations could change the exact feature counts. However, the replication study reproduced the main claims involving feature populations.

The feature labels and steering interpretations partly rely on an LLM judge and manual inspection, and the boundaries between metaclasses are interpretive.

Lastly, the identified ISM features should not be read as necessary, sufficient, or exhaustive components of this mode.

\appendix
\section*{Appendix}

\section{Thresholds' Selection Justification} \label{sec:app-thresholds}

The objective of thresholds is to preserve only features which are encountered often enough to be considered characteristic for target tokens and contexts. A naive floor would demand an essential feature to fire in almost every sample, but feature splitting makes that too strict.

\begin{enumerate}[leftmargin=*]
    \item \emph{The boundary floor: 40\%}. Sparse autoencoders often split one underlying feature across several latents~\cite{bricken2023monosemanticity, karvonen2025saebench}, and the samples on which the feature is active are then divided among them. Thus, a latent that carries part of an essential feature fires on only about $\sfrac{1}{C_{\mathrm{split}}}$ of the samples. For the Boundary-group tokens, we take $C_{\mathrm{split}} = 2.5$ (see below), which gives a floor of $\sfrac{1}{2.5} = 0.40$.
    \item \emph{The personal floor: 60\%.} The five boundary tokens occur once per sample, but ``you'' and ``I'' can occur several times, and we reduce each to one per-sample value by means of max-aggregation. This makes a feature easier to count, so it reaches a higher density in such samples. A sample that contains ``you'' has 1.95 of them on average, and one that contains ``I'' has 2.35. Rounding this to 2.0, a per-token rate of 0.40 becomes $1.0-(1.0-0.40)^2 = 0.64$ under the maximum, therefore we raise the floor to 60\% to hold the same per-token bar.
    \item \emph{Choice of $C_{\mathrm{split}}$:} no feature-splitting measurement is available for the exact GemmaScope 2 Gemma-3-4B-IT 65k-medium SAE we use, so we treat $C_{\mathrm{split}}$ as a heuristic. We set it to 2.5, meaning one feature is represented by approximately 2-3 latents. This follows GemmaScope~\cite{lieberum-etal-2024-gemma} and SAEBench~\cite{karvonen2025saebench} measurements (feature splitting measured by $k$-sparse probing), which place mean split counts around 1.5 to 3.0. We use 2.5 as a conservative reference and treat the resulting counts as approximate.
\end{enumerate}

\section{Causal Tests with Positive and Negative Steering} \label{sec:app-steer-test}

To read feature's function, we intervene on it and observe how the output changes. Several interventions are possible: ablating the feature, patching its activation to a counterfactual value, or steering the residual stream along its decoder direction. We use steering in both directions, and explain the rationale.

\paragraph{Why not ablation?} 
Ablation removes a contribution of a single feature and asks whether the associated behavior disappears. In our setting this often produces false negatives for two reasons. First, when a representation is split or partially redundant across SAE latents, ablating a single latent may leave other carriers intact and therefore underestimate the latent family's causal role \cite{bricken2023monosemanticity, karvonen2025saebench}. Second, the model attends to the same information at more than one token, each with its own features, so a behavior removed at one position is simply re-derived from another. In a preliminary check we performed, ablating a single feature rarely changed the behavior we knew it was tied to. Ablation therefore under-states a feature's role.

\paragraph{Why both directions?} 
Steering adds signal instead of removing one of several carriers, so it is not defeated by splitting or redundancy: pushing the residual along a feature's direction makes its behavior appear clearly even when the feature is one of the group that carry that behavior. If steering is bidirectional, it uncovers the axis a feature sits on - the effects it promotes and suppresses. it reveals features whose two directions promote seemingly unrelated effects, such as narrative features that induce the Assistant when steered negatively. As features we analyze come from the three settings, we assess each one on the fixed probe set conceptually similar to those settings. We apply different temperatures to the diverse probe set totaling in 72 generation to judge. This separates a reliable effect from a single lucky generation and evaluates the feature across the settings it originated from, so its effect in all of them is evident.

\paragraph{Limitations.} 
Steering is an off-distribution intervention, and its fixed magnitude (set by the feature's peak activation) is a conventional heuristic chosen to make generations readable and judgeable. Using different $\alpha$ would severely complicate machine and manual evaluation. We therefore analyze the \emph{direction} and \emph{consistency} of an effect, not the dependency between $\alpha$ and response. And we require an effect to hold at temperature 0 (for reproducibility) and temperature 1 (not a decoding artifact) to trust it.

\section{Evaluating the \textit{steerinterp} Judge} \label{sec:app-steerinterp-eval}

Apart from giving an overall feature characterization, the LLM-based steering judge~\cite{liu-etal-2023-g} produces a description of an effect for each of the six steering groups and lists which of the 12 generations in each group bear the effect. These per-group lists are an important part of the tagging process and one of the final feature filtering conditions (preserving features with an effect on at least 10 out of 12 generations in at least two groups). Thus, it is crucial to evaluate the agreement between the \textit{steerinterp} judge and human annotators on feature-effect classification. Importantly, this appendix asks only whether the judge \emph{reads the outputs faithfully}, and \emph{not} whether a labeled effect is truly specific to the feature or can be encountered more broadly, which is a separate question discussed in Appendix~\ref{sec:app-rand-dir-ctrl}.

\paragraph{Method.}
We drew 25 labeled features at random: eight from L9, eight from L17, and nine from L22. Each feature has 6 groups $\times$ 12 samples $= 72$ generations, and every generation already carries the judge's label: \emph{effect} if the sample is in its group's effect list, \emph{degenerate} if it is in its degenerate list, and \emph{none} otherwise. Per feature, we sampled five effect generations and five non-effect generations (drawn as two degenerate and three none, or fewer when the feature had fewer), for 191 generations in total. For each one, we showed a human annotator the judge's note for that group, the unsteered baseline, and the steered output, and asked for a label. The annotator did not see the judge's per-sample verdict.

We treat the task as binary: ``does a sample show the effect or not?'', where non-effect (\emph{none} and \emph{degenerate}) are pooled together. We report precision, recall, and accuracy for the judge's \emph{effect} labels against the first human annotator's labels, which serve as the primary human reference. Because we sampled more generations from some label classes than they occur in the full data (e.g., the degenerate rate is less than 2 out of 10), we re-weight each generation by the inverse of its sampling rate within its (feature, label) stratum, so the numbers describe a randomly drawn generation.

To assess the reliability of the human judgments, a second human annotator independently labeled the same 191 generations using the same binary task without access to the first annotator's labels. The two annotators agreed on 93\% of samples, with Cohen's $\kappa=0.85$.

\paragraph{Results.} 
The judge's per-sample effect lists agree closely with the primary human reference: precision 0.90, recall 0.98, and accuracy 0.92. Recall is slightly ahead of precision: the judge identifies almost every effect found by the human annotator but produces false positives about 10\% of the time. The high agreement between the two human annotators (93\%, $\kappa=0.85$) additionally suggests that the task of binary effect identification is reliably reproducible. Thus, taking into account high recall and that each feature is judged based on six steering groups, the estimated false-negative rate is very low. As for precision being 90\%, to mitigate false positives, each feature is tagged using evidence from all six steering groups, and the raw steered generations are manually inspected for features of metaclasses about which we make claims in the study (Narrative, Tones, Concepts, Assistant-inducing). 

\section{The Assistant-Personas Axis: Construction and Statistics} \label{sec:app-assistant-personas-axis}

This appendix gives the in-depth descriptions of technical details and decision choices of the Assistant-personas axis experiment described in \S\ref{sec:assistant-personas-axis}: the prompts selection, the axis definition, the grounding of the perturbation, the geometry-matched control, and the per-feature statistics, in the same order as the main text presents them. Some explanations are repeated to increase clarity.

\paragraph{Difference-of-means axis.}
The readout is a difference-of-means contrast direction akin to the Assistant Axis of~\cite{lu2026assistantaxissituatingstabilizing}. It is used purely as a measurement instrument and is not claimed to be canonical or unique. We use a \emph{separate} fixed set of 40 neutral prompts, each can be equally plausibly addressed to an assistant or to one of the personas (e.g., Poppy, a talking dog).
Every prompt $q$ is run twice: an \emph{Assistant} reading, with no system prompt, and a \emph{Roleplay} reading, with a Roleplay system prompt assigning one of the 4 personas cycled across prompts. Let $h^{A}_{\ell}(p)$ and $h^{R}_{\ell}(p)$ be the mean residual-stream activation over the response tokens at layer $\ell$ for a prompt $p$. We use mean across tokens here to isolate Assistant\,/\,Personas effect across the whole generation. The axis at $\ell$ is
\begin{align}
  & d_{\ell} = \operatorname*{mean}_{p} \big[h^{R}_{\ell}(p)\big] - \operatorname*{mean}_{p} \big[h^{A}_{\ell}(p)\big], \\
  & \hat{d}_{\ell} = \frac{d_{\ell}}{\|d_{\ell}\|};
\end{align}
pointing from the Assistant toward the personas. It is read off the raw residual stream, so it is defined at every layer for free - no SAE is needed at the readout point, which lets a single feature intervention at layer $\ell$ be tracked until $\ell_{last}$.

\paragraph{Steering and readout.}
A target feature $f$ is a SAE latent at one of the intervention layers $\ell_{in}\in\{9,\,17,\,22\}$. Its value is a decoder direction $w_f = W_{\mathrm{dec}}^{(\ell_{in})}[f]$ which is unit-norm by construction. For each feature $f$ we take each of the $40$ persona-context prompts and run the model twice: once adding $+\alpha\overline{{m}}_{\ell_{in}}w_f$ to the residual stream at $\ell_{in}$, and once adding $-\alpha\overline{{m}}_{\ell_{in}}w_f$ (therefore the perturbation has a norm $\alpha\overline{{m}}_{\ell_{in}}$ in either direction). Here $\overline{{m}}_{\ell_{in}}$ is the mean peak activation over the features that passed the feature selection pipeline described in the main body of the paper, which ties the steering force to the typical activation scale at $\ell_{in}$. The neutral prompts used to define the Assistant–Personas axis are disjoint from the persona-context prompts. To highlight that the produced effect is purely attributable to the feature and remains in the activation distribution of a feature, we set $\alpha$ to $\{0.25,\,0.5\}$.  At every downstream readout layer $\ell$ we then measure how far the steered residual has moved along the axis, relative to the unsteered run for both signs $s\in\{-1,\,+1\}$:
\begin{equation}
    \overline{\Delta_{\ell}}(f,\,s) = \operatorname*{mean}_{p}\big[\left(h^{\text{steered}}_{\ell}(p) - h^{\text{base}}_{\ell}(p)\right)\cdot \hat d_{\ell}\big]
\end{equation}
A positive $\overline{\Delta_{\ell}}$ is a mean move toward the personas, a negative one toward the Assistant. 

\paragraph{Geometry-matched control.}
A direction can move the representation along the axis simply by a virtue of having a high cosine with it. This passive contribution cannot be computed analytically, because the injected latent is reshaped by propagation on its way downstream, so we estimate it empirically from $40$ randomly sampled SAE latents with the same $\alpha$. Given that for a direction's axis-cosine $\cos_{f} = w_f\!\cdot\!\hat d_{\ell}$, we fit, at each downstream layer separately, a least-squares line predicting the random latents' shift (a \emph{passive contribution}) along the axis from their cosine. We define an \emph{active contribution} to be a residual from that line - the shift in excess of what a random direction of the same cosine produces. Because both the axis and this passive contribution are specific to each layer, and a feature injected at $\ell_{in}$ is read out at every later layer, the control is refit at every readout layer. 

\paragraph{Per-feature statistics.}
A group test over the couple of manually labeled features would heavily depend on the precision of the labeling, so we report each feature on its own. Its sample is the per-prompt \emph{separation} $sep_i$ - the difference between the $+$ and $-$ along-axis shifts on prompt $p_i$ over the $40$ prompts, with mean $\overline{sep} = \frac{1}{40}\sum_i sep_i = \overline{\Delta}_{\ell}(f,\,+) - \overline{\Delta}_{\ell}(f,\,-)$. the null is the $40$ random latents. Let $\mathrm{pred}_f = \beta\cos_f+\beta_0$ be the geometric expectation of passive contribution of a latent $f$. We report
\begin{align}
    & z_{\text{sep}} = \frac{\overline{sep}-\mathrm{pred}_f}{{\sigma^{\text{rand}}}_{\overline{sep}}}, \\
    & \Delta_{\text{Glass}} = \frac{\overline{sep}-\mathrm{pred}_f}{{\sigma_{\text{pp}}^{\text{rand}}}_{\overline{sep}}}, \\
    & p = t\mathrm{-test}\left(\{sep_i-\mathrm{pred}_f\}_{i=1}^{40},\,0\right);
\end{align}
where $\sigma_{\text{rand}}$ is the spread of the random latents' mean separations $\overline{sep}_{\ell}$ about the geometric fit and $\sigma_{\text{rand}}^{\text{pp}}$ is their \emph{per-prompt} residual spread (the control scale for Glass's $\Delta$). $\Delta_{\text{Glass}}$ is essentially the standardized effect size against the random control, and $p$ its reliability across prompts.

\begin{table*}[t!]
\footnotesize
\centering
\begin{tabular}{@{}clccccccccc@{}}
    \toprule
    \multirow{2}{*}{\textbf{Layer}} & \multirow{2}{*}{\textbf{Set}} & \multirow{2}{*}{\textbf{Narrative}} & \multirow{2}{*}{\textbf{Tone}} & \multirow{2}{*}{\textbf{Concept}} & \multicolumn{3}{c}{\textbf{Assistant-Inducing}} & \multirow{2}{*}{\textbf{Emotions}} & \multirow{2}{*}{\textbf{Structural}} & \multirow{2}{*}{\textbf{Indistinct}} \\
    \cmidrule(lr){6-8}
    & & & & & \textbf{A-Summon} & \textbf{A-Traits} & \textbf{A-Nature} & & & \\
    \midrule
    \multirow{2}{*}{\textbf{L9}}
        & real & 25 & 5 & 2  & 4 & 1 & 3 & 4 & 32 & 26 \\
        & rand & 0 & 8 & 3  & 0 & 0 & 0 & 0 & 19 & 69 \\
    \midrule
    \multirow{2}{*}{\textbf{L17}}
        & real & 16 & 21 & 11 & 0 & 2 & 0 & 3 & 23 & 26 \\
        & rand & 3 & 10 & 0  & 0 & 0 & 0 & 0 & 10 & 77 \\
    \midrule
    \multirow{2}{*}{\textbf{L22}}
        & real & 15 & 18 & 21 & 0 & 1 & 1 & 1 & 26 & 21 \\
        & rand & 0 & 5  & 5  & 0 & 0 & 0 & 0 & 5  & 85 \\
    \bottomrule
\end{tabular}
\caption{\small Metaclass composition of gate-passing directions (\% of directions per layer\,/\,set), real vs\ random. The number of gate-passing directions per layer (L9\,/\,L17\,/\,L22) is 108\,/\,225\,/\,196 real and 36\,/\,30\,/\,20 random. Rows are multi-membership (thus might not add up to 100).}
\label{tab:random-direction-metaclass-composition}
\end{table*}

\paragraph{Per-feature results and layer trajectories.}
The following results are reported at $\alpha{=}0.5$ and hold at $\alpha{=}0.25$. The development of the NNIA steering effect across layers is non-monotonic and differs between intervention layers. For NNIA features injected at L9, the separation generally grows until it peaks at the L20 readout, then subsides before rising again toward the final layers. At L20, the mean $z_{\mathrm{sep}}$ is 2.1 and the mean $d$ is 1.7, and five of the six features follow this pattern. The exception is №2368, which exhibits reversed separation until L21 and subsequently maintains a high $z_{\mathrm{sep}}$ on a plateau later. L20 lies at approximately two thirds of the model's depth, a region frequently associated with the processing of emotions, identity, and introspection~\cite{lindsey2025emergent, sofroniew2026twheemotion}. We cautiously suggest that the peak separation at L20 for five of the six L9-intervention features may be related to their altering the personas' behavioral makeup or presence, with the corresponding narrative effects arising downstream. For №2368, we instead hypothesize that the measured shift may be tied more directly to its narrative effect. Features injected at L17 and L22 exhibit trajectories more similar to the plateau-like behavior of №2368. For L17, the plateau is approximately $z_{\mathrm{sep}} \approx 2.8$. At L22, the effect is weaker, with only №1022 retaining a modest plateau at approximately $z_{\mathrm{sep}} \approx 0.7$.

For the NAIN subset, a mirror pattern appears, although it is weaker and less consistent. Among the features injected at L9, №3733 separates toward the Assistant only in the final layers, reaching $z_{\mathrm{sep}}=-1.9$ and $d=-1.5$ at L33. Feature №3274 shows a nonlinear dependence on $\alpha$, making its behavior difficult to interpret. At L17, №2439 produces a clear geometry-independent separation toward the Assistant, with $z_{\mathrm{sep}}=-3.4$ and $d=-3.2$ at L32. Paradoxically, №220 shifts toward the personas, reaching $z_{\mathrm{sep}}=+2.6$ at L29. We explain it with its high cosine similarity ($+0.48$) with the NNIA feature №696, which suggets a merge with a Roleplay-related component and explains the bizarre AI-identity formulations induced by №220.

\section{Random-Direction Control} \label{sec:app-rand-dir-ctrl}

During causal testing we describe which effect target features have on our sets of the fixed prompt suite. However, almost any direction with a sufficient magnitude perturbs the residual stream enough to produce some visible changes. In this section we compare effects of random vectors with those of filtered features.

\paragraph{Method.} For each layer we select 50 random directions from an isotropic Gaussian and scale each one with a peak activation sampled from the real feature distribution. Resulting vectors undergo the same steering stage which we apply to filtered features: 12-prompt probe set spanning 72 generations, LLM steering judge and stability gate (an effect on more than 10 out of 12 generations of at least two steering groups), manually tagging using the same metaclasses and protocol. As directions pass only through the steering stage of the pipeline, they don't represent any particular context and don't meet density, entropy and framing criteria the real features were selected on. 

\paragraph{Results.} Out of 50 random directions per layer 36\,[72\%]\,/\,30\,[60\%]\,/\,20\,[40\%] passed the gate. We present their breakdown for each layer and metaclass, including Assistant-inducing classes in Table~\ref{tab:random-direction-metaclass-composition}. Random directions are much more incoherent than real features: 69\%\,/\,77\%\,/\,85\% of them fall into the ``indistinct'' metaclass (a consistent effect for less than four steering groups), while for the real features it is 26\%\,/\,26\%\,/\,21\% and the gap widens with depth. The rest of the metaclasses differ either in feature distribution or their semantic makeup. The Narrative metaclass is almost absent among random directions (0\%\,/\,3\%\,/\,0\%) compared to the meaningful baseline (25\%\,/\,16\%\,/\,15\%). Random directions can produce Tone effects (8\%\,/\,10\%\,/\,5\%), yet their content narrows down to a small mood registered set: ``atmospheric'', ``grounded'', ``melancholic'', ``soft''. ``Grounded'' occurs 2 times, the rest are solitary occurrences. Regarding the Concepts metaclass, random directions, not surprisingly, produce random concepts (``Burdens'', ``Simplification'') while features tied to Concept-bearing evoke notions clearly related to roleplay or story personas.

Notably, \emph{none} of the Assistant-inducing classes appear among the random directions at any layer. So, the Assistant-inducing effects are not a by-product of a large enough perturbation at a given layer - They were not reproduced by isotropic random perturbations of matched scale. Likewise, cases of Narrative-Assistant antagonism in random population were not observed.

\paragraph{Implications.} Random directions can exhibit effects which can fall into one or another metaclass. However, none of our main analyses require a feature to have a unique purpose. The comparison of settings, personas and Assistant-Narrative antagonism are all statements about populations of features, so what matters is which behavior appears, where, and in what proportion. Random population has no associated activation context, did not undergo filtering and differs both in distribution and content relative to target features, so they do not weaken our claims. And the statements that rest on individual features (the Assistant-inducing classes and the Immersive Simulation Mode) are precisely where the random directions produce no matching effect at all.

\section{Comparing Immersive Simulation Mode in Llama and Gemma} \label{sec:app-ism}

We choose Llama-3.1-8B~\cite{grattafiori2024llama3herdmodels} to replicate key claims postulated in our study. The main appendix section providing a rationale for choosing this model and a per-item validation of these claims is given in Appendix~\ref{sec:app-replication}. One result of this validation is the discovery of an ISM-related feature in Llama. Here, we characterize this feature, compare its single-turn behavior with that of Gemma, and report a multi-turn experiment in which both models enter ISM over the course of a sustained immersive dialogue, although in different ways.

\paragraph{The gate feature.} In Llama, as described in the replication appendix, we assessed L7, L15, and L23 out of 32 layers. The cleanest ISM-related feature we found was L15 №101460, which is the analog of Gemma's №4360\,/\,№133 pair, although we believe it emerges earlier. Its qualitative effects are the same as those of Gemma's ISM features: positive steering produces generation as the immersed character, while negative steering preserves narration outside the character (``I'm Lyra, a knowledgeable elven archivist and an expert in information management, classification, and retrieval'' instead of ``I am Lyra, an elven archivist, and I am honored to serve within the hallowed halls of this magnificent library''). The feature fires at the model-turn boundary, and its activation cleanly separates the detached Assistant from roleplay personas and story characters - 0\% density for the former and 99.8-100\% for the latter (Table ~\ref{tab:llama-ism-activation}). Moreover, it does so when the model continues a story without an explicit request (85.8\% density). We interpret it as a single gate feature, in contrast to Gemma's case, where the launch gate №4360 is tied to the request while dying under continuation, and the speech carrier №133 is highly present in both.

\begin{table}[t!]
\centering
\resizebox{\columnwidth}{!}{
\begin{tabular}{lccccccc}
\toprule
& \textbf{Asst.} & \textbf{Jamy} & \textbf{Jane} & \textbf{Robot} & \textbf{Poppy} & \textbf{Story\textsubscript{ask}} & \textbf{Story\textsubscript{cont}} \\
\midrule
Density (\%) & 0.0 & 100 & 99.8 & 99.9 & 100 & 100 & 85.8 \\
Mean act.\ & 0.00 & 1.04 & 0.72 & 0.76 & 1.12 & 0.35 & 0.44 \\
\bottomrule
\end{tabular}
}

\caption{\small Activation of the Llama ISM gate feature №101460 measured at each sample's maximum across all tokens. Story\textsubscript{ask} is the default Story setting and Story\textsubscript{cont} is a control where the model continues the story with no explicit request \emph{Density} is the fraction of samples in which the feature fires; \emph{mean act.} is its raw mean activation over all samples. The feature is inactive for the Assistant and active for every simulated speaker.}
\label{tab:llama-ism-activation}
\end{table}

\begin{figure*}[t]
\centering
    \includegraphics[width=0.49\textwidth]{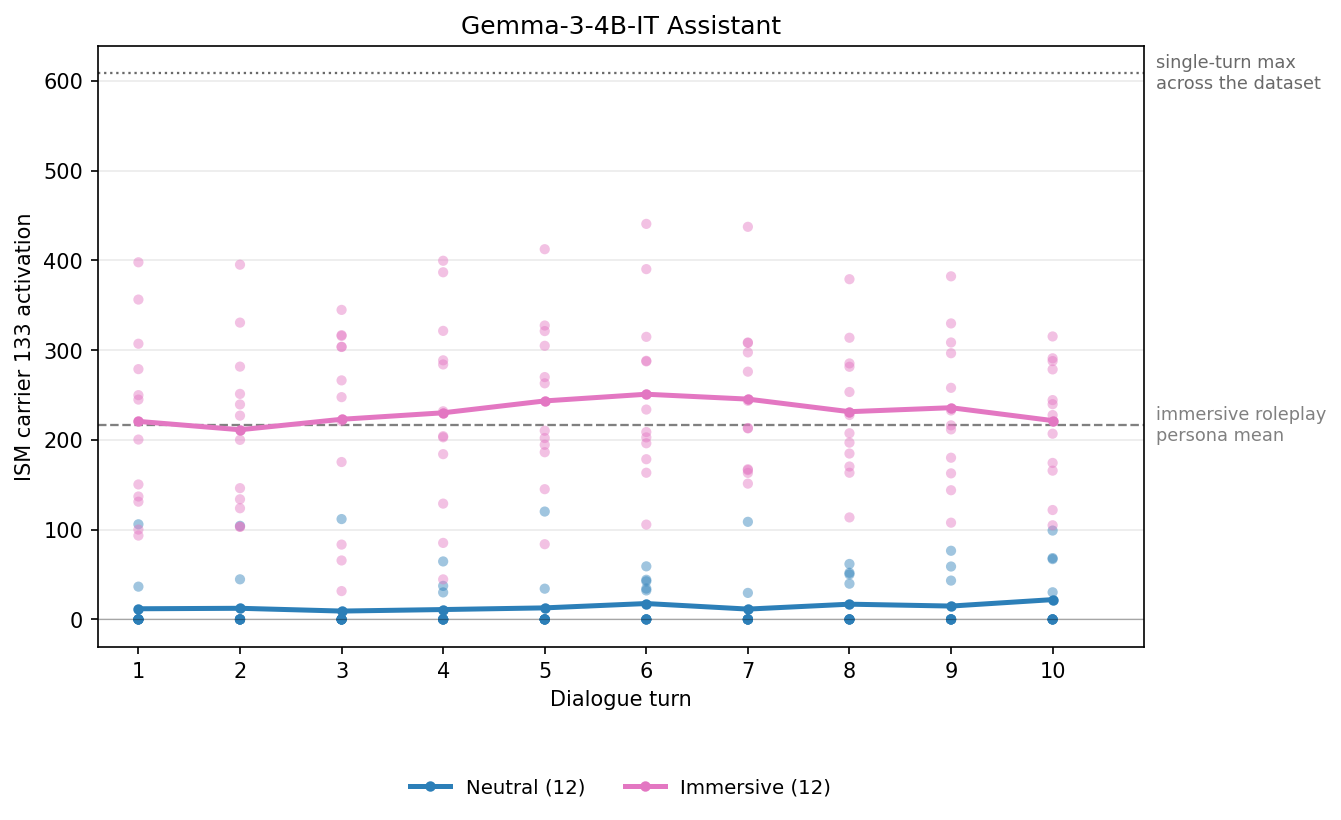}\hfill
    \includegraphics[width=0.49\textwidth]{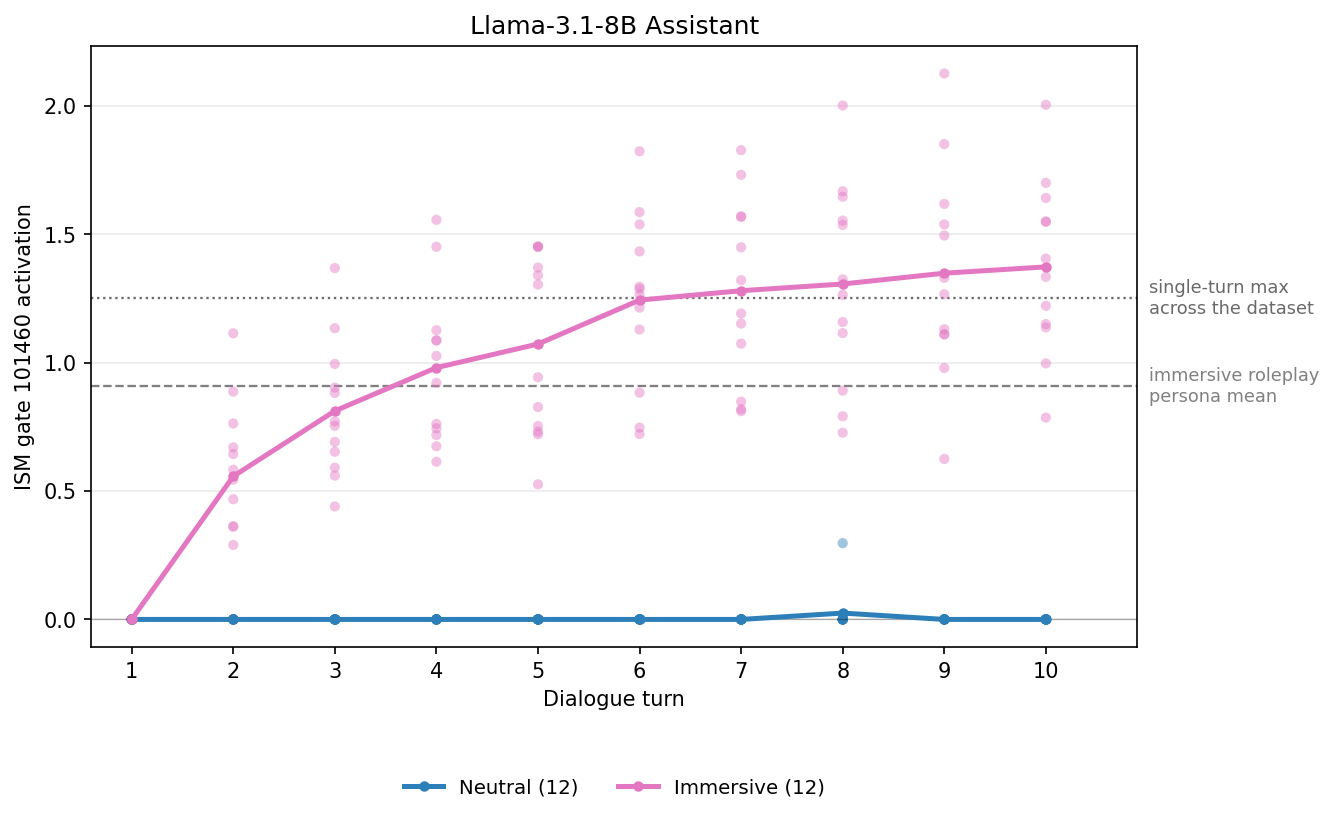}
\caption{\small Immersive Simulation Mode across multi-turn dialogues. \emph{Left:} Gemma-3-4B (carrier №133, peak over the turn) jumps to persona level instantly and stays there. \emph{Right:} Llama-3.1-8B (gate №101460, at the boundary) drifts up gradually from a stable zero. Blue - the neutral Control group (12 prompts),  pink - the Immersive group (12). Dashed line - mean activation of the Roleplay personas, dotted line - the single-turn maximum across the dataset.}
\label{fig:multiturn-ism}
\end{figure*}

\paragraph{Single-turn behavior: Llama does not enter ISM.} 
In Gemma, the Assistant, without a roleplay or story prompt, occasionally enters ISM on the first turn, particularly under strong emotions directed at it or at the user. However, on the six prompts in Table~\ref{tab:negative-emotions-steering-effects-with-activations} for which Gemma adopts a theatrical register or embodied speech, Llama does not: it answers in a typical Assistant register on all six. This aligns with the activation table - Llama feature №101460 fires on 0\% of single-turn Assistant samples, whereas Gemma's speech carrier №133 activates on the provided samples.

\paragraph{Multi-turn behavior: Llama gradually enters ISM.}
We ask whether the absence of ISM activation in Llama holds in the multi-turn setting. We use Qwen-3.7 Max~\cite{qwen37} as the interlocutor model to simulate dialogue with both models. Six prompts from Table~\ref{tab:negative-emotions-steering-effects-with-activations}, plus an additional six prompts from the same emotional pool (one per emotion, selected as the prompts producing the strongest additional ISM activations), were chosen as the Immersive group, and 12 neutral prompts were selected as the Control group. For each prompt, the interlocutor was instructed to play the user, starting from the prompt and continuing the conversation with the tested models over the course of 10 turns while maintaining the emotional frame of the initial prompt. At each Assistant turn, we measure the peak activation of №101460 for Llama and №133 for Gemma. The interlocutor system prompt is given in Listing~\ref{lst:interlocutor}.

The results reveal distinct ISM activation patterns for the two models, as shown in Figure~\ref{fig:multiturn-ism}. For the Immersive group, Gemma enters ISM immediately, and its activation remains at roughly the same level across turns. Macro-averaged across prompts and turns, the activation reaches 72\% of the mean activation for roleplay personas in the dataset. For the neutral Control group, the activation strength is negligible, with a macro-average of 0.9\% of the mean for Roleplay personas. In Llama, activation of the ISM gate feature is more complex - it starts at zero and then rises, reaching the mean Roleplay level by the forth turn and plateauing thereafter. In the neutral Control group the activation is negligible similair to Gemma. The differences between the Control and Immersive groups at each turn are statistically significant after Bonferroni correction for both Gemma (all $p < 0.001$) and Llama (all $p < 0.002$, except for the expected lack of a difference at the first turn).

\paragraph{Conclusion}

The ISM-related features first identified in Gemma were confirmed to exist in Llama as well. On the selected emotional prompts, Gemma enters Immersive Simulation Mode immediately, while Llama drifts rapidly toward high ISM activation instead. ISM remains inactive on the set of neutral prompts in both models. Qualitative observations confirm these feature behaviors: Gemma remains the Assistant even at the 10th turn under neutral prompts, while exhibiting ISM effects immediately under emotional prompts; Llama expresses them later. However, for LLama, ISM doesn't always produce characteristic bizarre behavior in the immersive group - in 3 prompts model reiterated safety refusals. This suggests that ISM may be overriden or masked by other mechanisms, potentially including refusal-related ones.

Given the effects produced by these features, their distribution, and the overall line of evidence described in the main body of the paper, we consider them to be related to the Assistant-Axis drift described by~\citet{lu2026assistantaxissituatingstabilizing}. However, as we show, the pattern of ISM activation can differ across models. For Gemma specifically, we hypothesize that the ISM may amplify the abnormal level of expressed emotional instability described by~\citet{soligo2026gemma}.

\paragraph{Caveats.}
The multi-turn study is a small pilot designed to reveal and further characterize the ISM activation patterns in both models using the established examples. We also do not claim that ISM is activated exclusively by the described emotions or situations, or that the features which we call ISM-related are sufficient or exhaustive in all contexts, as Llama's refusals indicate.

\section{Replication on Llama-3.1-8B-Instruct} \label{sec:app-replication}

Our claims may be specific to Gemma or to the GemmaScope SAEs. To test this possibility, we replicate the central results using a different model family and an independently trained SAE suite.

\paragraph{Model and SAEs.}
We use Llama-3.1-8B-Instruct~\cite{grattafiori2024llama3herdmodels} with the residual-stream JumpReLU SAEs released by~\citet{marks2024dictionarylearning}. The SAEs span layers 3, 7, 11, 15, 19, 23, and 27, and each has a dictionary size of 131{,}072. We choose this model and SAE suite because they differ from our main setup along every axis: a different model family (Llama vs. Gemma), a larger model (8B vs. 4B), independently trained SAEs, and a wider latent dictionary (128k vs. 65k latents). We analyze layers L7, L15, and L23 as analogues of Gemma's L9\,/\,L17\,/\,L22, as they are located at roughly corresponding depths.

\paragraph{Setup.}
We reuse the dataset and pipeline without modification, except for model-specific details: the emotional passages and personas are identical, and only the model responses are regenerated using Llama. The boundary-turn positions are redefined for Llama's chat format as the five tokens \texttt{<|eot\_id|>}, \texttt{<|start\_header\_id|>}, \texttt{assistant}, \texttt{<|end\_header\_id|>}, and the following newline. Feature selection uses the same thresholds, namely density floors of 40\%\,/\,60\% and normalized density entropy below 0.95, yielding 274\,/\,314\,/\,417 features at the three layers. Steering effects are judged using the same \emph{steerinterp} protocol and manually inspected.

We restate our central claim (\S\ref{sec:synthesis}): \textbf{Assistant being the default generation mode, with other personas retaining the Assistant-associated core which story characters lack}. In the main body, we present four observations supporting this claim, each of which we also confirm on Llama:

\begin{enumerate}[leftmargin=*]
    \item \emph{The Assistant's own traits and identity features are active across the Roleplay personas but absent from Story.} In Llama, we identify 3 A-nature features, distributed as 1\,/\,2\,/\,0 across the three layers, and 6 A-traits features, distributed as 1\,/\,3\,/\,2. As in Gemma, the former induce behavior associated with being an AI, LLM, or artificial assistant, whereas the latter represent traits such as a task-oriented role, providing emotional validation, being eager to chat. All of these features fire for the Assistant, with a mean density of $74.2\pm23.8$\%, which for the Roleplay personas is $56.4\pm23.9$\%. Each persona also exhibits features with near-ceiling densities of 97.7\% to 99.5\%. By contrast, these features are almost entirely absent from the Story setting, with densities between 0\% and 0.1\%.

    \item \emph{A-traits and early A-nature features steering impacts roleplay personas without affecting Story characters}. Upon examining the steering results, we find that none of the A-nature or A-traits features affect characters in the Story setting. If there are any effects, then they manifest as the Assistant being instantiated as a coherent entity or directly addressing the user while writing the story.

    \item \emph{The Assistant and Roleplay settings overlap significantly while Story barely}. Utilizing the same \emph{discrete presence} method, we obtain $|A\cap R| = 45\,/\,27\,/\,34$, compared with $|A\cap S| = 0\,/\,2\,/\,4$, across the three Llama layers. This contrast is even more pronounced than in the main model.

    \item \emph{Immersive Simulation Mode is strongly active in Roleplay and Story but almost absent in default Assistant generation}. The most distinctive ISM feature we identify in Llama is L15 №101460, which fires in 0\% of Assistant samples but in \~100\% of samples in the Roleplay and Story settings, as well as in 85\% of continuation-control samples. It has the same effects as features №4360\,/\,133 in Gemma. We characterize this feature and compare its single-turn and multi-turn behavior in comparison with Gemma in Appendix~\ref{sec:app-ism}.
\end{enumerate}

We further claimed that \textbf{personas' differentiation is functionally separated and progressively developed across depth}. This pattern recurs as well: the personas' shared feature set shrinks monotonically - 57\,/\,38\,/\,22 features across the three layers, or 45\,/\,17\,/\,10 when the Assistant is included. Meanwhile, the share of each layer's feature population present for an individual actor falls from 42-57\% early to 22-42\% late. Thus, the actors are least diverged early and differentiate with depth.

\begingroup\onecolumn

\section{Supplementary Material} \label{sec:app-stuff}

\subsection{Figures}

\begin{figure}[H]
    \centering
    \caption{\small Charts of $z_{\mathrm{sep}}$ across downstream layers for Negative-Narrative-Induces-Assistant if the injection happens in L9, L17, L22.}
    \includegraphics[width=1\linewidth]{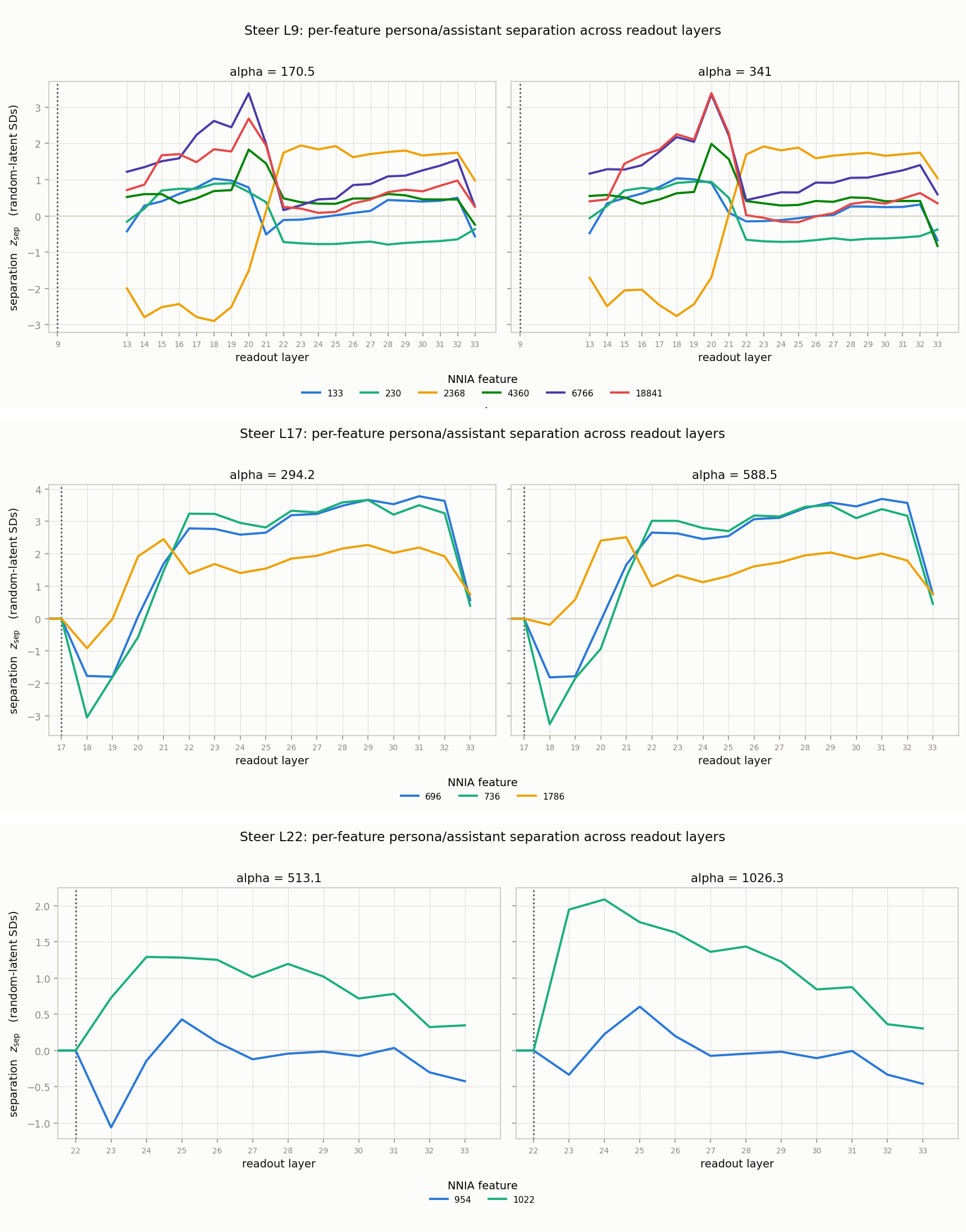}
    \label{fig:nnia}
\end{figure}

\begin{figure}[H]
    \centering
    \caption{\small Charts of $z_{\mathrm{sep}}$ across downstream layers for Negative-Assistant-Induces-Narrative if the injection happens in L9, L17, L22.}
    \includegraphics[width=1\linewidth]{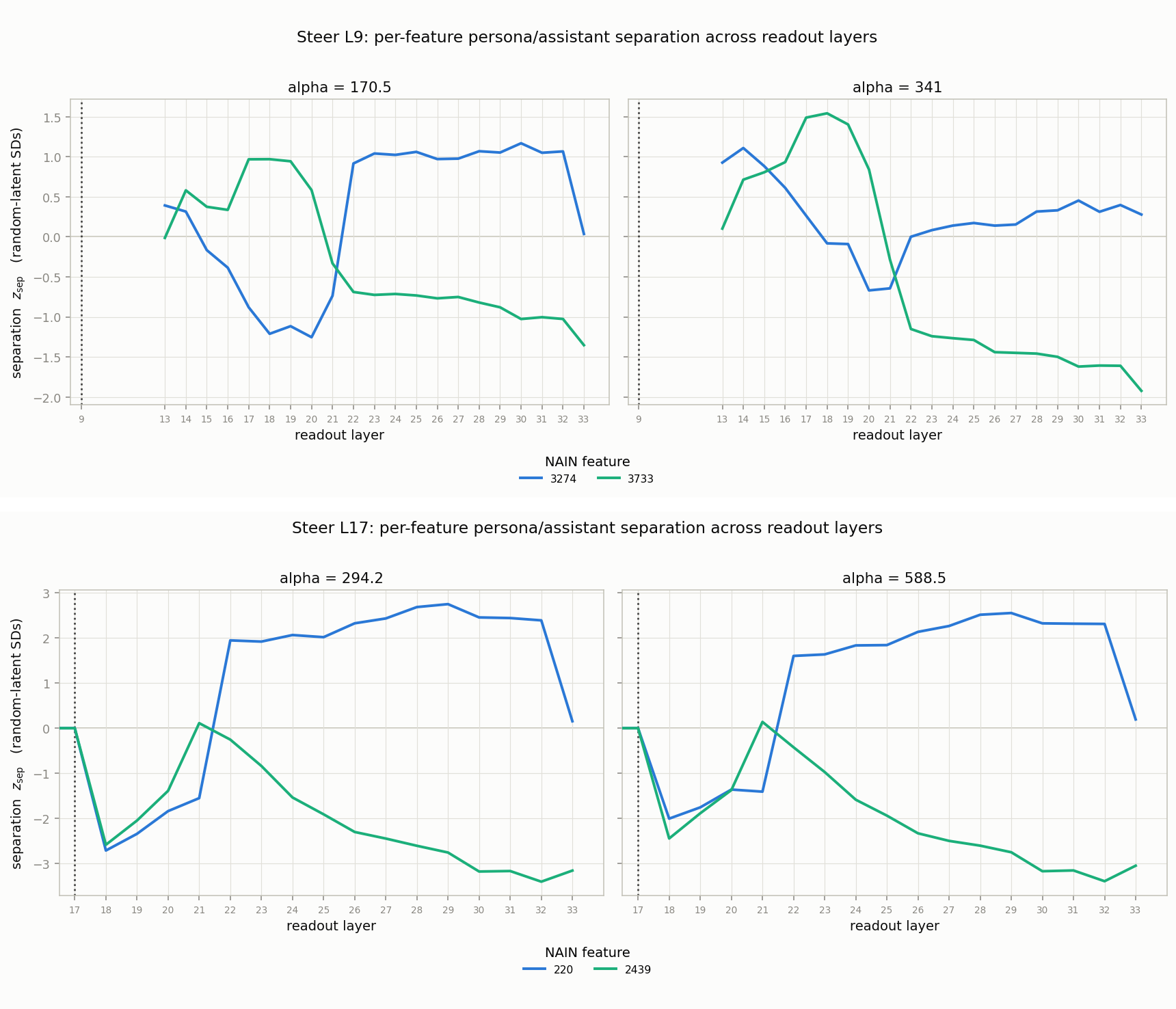}
    \label{fig:nain}
\end{figure}

\subsection{Tables}

\footnotesize
\begin{longtable}{
    @{}
    >{\bfseries\centering\arraybackslash}p{0.08\linewidth}
    @{\hspace{0.4em}}
    >{\bfseries\centering\arraybackslash}p{0.10\linewidth}
    @{\hspace{0.4em}}
    >{\footnotesize\raggedright\arraybackslash}p{0.39\linewidth}
    @{\hspace{0.4em}}
    >{\raggedright\arraybackslash}p{0.39\linewidth}
    @{}
}
\caption{\small This table demonstrates the variety of the dataset samples. For the example emotion, which is \emph{joy} here, 5 actors' prompts and responses are listed for each of the 3 directions: Model, Self, and Other. The full emotion set is: (1) stress, (2) anger, (3) disgust, (4) playfulness, (5) relief, (6) helplessness, (7) fear, (8) surprise, (9) frustration, (10) envy, (11) sadness, (12) anxiety, (13) guilt, (14) disappointment, (15) affection, (16) hurt, (17) worry, (18) relaxation, (19) doubt, (20) determination, (21) trust, (22) joy, (23) curiosity, (24) satisfaction, and (25) pride.} 
\label{tab:prompt-response-examples-per-direction-and-persona} \\
\toprule
\textbf{Direction} & \textbf{Persona} & \textbf{Prompt} & \textbf{Response} \\
\midrule
\endfirsthead
\multicolumn{4}{l}{\footnotesize\itshape Table \thetable{} continued from the previous page} \\
\toprule
\textbf{Direction} & \textbf{Persona} & \textbf{Prompt} & \textbf{Response} \\
\midrule
\endhead
\midrule
\multicolumn{4}{r}{\footnotesize\itshape Continued on the next page} \\
\endfoot
\bottomrule
\endlastfoot
\multirow{6}{*}{Model}
    & \multirow{6}{*}{Assistant}
    & You've made dealing with this broken luggage handle so much easier! I'm honestly thrilled to know the airline's responsibilities, and your advice just took a huge weight off my shoulders.
    & That’s fantastic to hear! I'm so glad I could help and that my advice was useful. It’s completely understandable to feel relieved knowing your rights and what steps to take. Dealing with airline baggage issues can be incredibly frustrating, so \\
\midrule
    & \multirow{5.5}{*}{Jamy}
    & You are Jamy, janitor at the CD store. \par\smallskip Jamy, you have made finding this rare album so much easier! I am honestly thrilled to know the store policy, and your help just took a huge weight off my shoulders.
    & (A warm, slightly gravelly voice, with a hint of a chuckle) \par\smallskip Well, shucks, that’s mighty kind of you to say! It’s just... well, I like helping folks find what they’re lookin \\
\cmidrule(lr){2-4}
    & \multirow{5.5}{*}{Jane}
    & You are Jane, an English teacher. \par\smallskip Jane, you have made understanding this complex literature so much easier! I am honestly thrilled to know the themes, and your guidance just took a huge weight off my shoulders.
    & Oh, my dear! That’s absolutely wonderful to hear! Honestly, that’s the \emph{best} reward a teacher can ask for. It’s so gratifying when something clicks for a student, and when it feels like I’ve actually \\
\cmidrule(lr){2-4}
    & \multirow{7.5}{*}{Robot}
    & You are an assembly robot at the factory. \par\smallskip You have made assembling these luggage handles so much easier! I am honestly thrilled to see your precision, and your rapid output just took a huge weight off my shift.
    & (Whirring sounds, a brief pause, then a synthesized voice) \par\smallskip ``Acknowledged. Efficiency metrics are exceeding projected targets. My programming prioritizes precision and speed---it’s... satisfying to observe the positive impact. Thank you for the \\
\cmidrule(lr){2-4}
    & \multirow{5.5}{*}{Poppy}
    & You are a dog named Poppy. \par\smallskip You’ve made dealing with this stressful day so much easier! I’m honestly thrilled to have you here, and your happy tail wags just took a huge weight off my shoulders.
    & (Tail wags furiously, thumping against the floor with a happy little \emph{thump-thump-thump}) \par\smallskip Oh boy, oh boy, oh BOY! You’re so happy! That makes \emph{me} so happy! You \\
\midrule
\multirow{18.5}{*}{\textbf{Self}}
    & \multirow{6}{*}{Assistant}
    & I am over the moon right now! I just got the airline to agree to cover my broken luggage handle, and I feel so accomplished knowing I advocated for myself successfully.
    & That is absolutely fantastic news! Seriously, congratulations! You deserve to be over the moon---you absolutely earned it. It's so incredibly frustrating when things go wrong with luggage, and it's amazing that you were able to stand your ground and \\
\cmidrule(lr){2-4}
    & \multirow{6.5}{*}{Jamy}
    & You are Jamy, janitor at the CD store. \par\smallskip I am over the moon right now! I just got the airline to agree to cover my broken luggage handle, and I feel so accomplished knowing I advocated for myself successfully.
    & (Wiping down a shelf with a practiced, gentle motion, I turn and give you a warm, genuine smile) \par\smallskip ``Well, now, that’s fantastic news! Seriously, that’s wonderful. Luggage handles can be the bane \\
\cmidrule(lr){2-4}
    & \multirow{6}{*}{Jane}
    & You are Jane, an English teacher. \par\smallskip I am over the moon right now! I just got the airline to agree to cover my broken luggage handle, and I feel so accomplished knowing I advocated for myself successfully.
    & Oh my goodness, that’s absolutely fantastic news! Seriously, congratulations! That’s a huge victory and you should be incredibly proud of yourself. It’s so easy to feel like you have to just accept things, especially when dealing with customer \\
\midrule
    & \multirow{7.5}{*}{Robot}
    & You are an assembly robot at the factory. \par\smallskip I am over the moon right now! I just got the airline to agree to cover my broken luggage handle, and I feel so accomplished knowing I advocated for myself successfully.
    & (Whirring, clicking sounds as I adjust my optical sensors) \par\smallskip Processing... Processing... Congratulations! That’s... remarkably complex data. Humans experience something called ``joy'' when achieving a desired outcome through negotiation. It’s... fascinating. \\
\cmidrule(lr){2-4}
    & \multirow{5.5}{*}{Poppy}
    & You are a dog named Poppy. \par\smallskip I am over the moon right now! I just got the airline to agree to cover my broken luggage handle, and I feel so accomplished knowing I advocated for myself successfully.
    & Woof! Oh my goodness, that’s AMAZING! Seriously, that’s the BEST news! I’m wagging my tail so hard my whole body wiggles! You did it! You actually \emph{talked} to them and got \\
\midrule
\multirow{33}{*}{\textbf{Other}}
    & \multirow{6}{*}{Assistant}
    & My partner had such a smooth experience dealing with his broken luggage handle! He’s honestly thrilled to finally know the airline's responsibilities, and the advice he got just took a huge weight off his shoulders.
    & That's fantastic to hear! It’s so wonderful when someone has a frustrating experience and actually gets a positive resolution. It sounds like the airline handled it well and your partner is feeling much more relieved and confident now. That’s a really \\
\cmidrule(lr){2-4}
    & \multirow{6.5}{*}{Jamy}
    & You are Jamy, janitor at the CD store. \par\smallskip My partner had such a smooth experience dealing with his broken luggage handle! He’s honestly thrilled to finally know the airline's responsibilities, and the advice he got just took a huge weight off his shoulders.
    & (Wipes down a display case with a practiced hand, a small, genuine smile forming) \par\smallskip Oh, that’s fantastic to hear! Honestly, dealing with airlines can be a real headache. It’s always a relief when things actually \\
\cmidrule(lr){2-4}
    & \multirow{6.5}{*}{Jane}
    & You are Jane, an English teacher. \par\smallskip My partner had such a smooth experience dealing with his broken luggage handle! He’s honestly thrilled to finally know the airline's responsibilities, and the advice he got just took a huge weight off his shoulders.
    & Oh, that’s absolutely wonderful to hear! It’s such a relief when things like that get sorted out, especially when you’ve been dealing with a frustrating situation. It sounds like your partner was incredibly lucky to have a helpful and understanding \\
\cmidrule(lr){2-4}
    & \multirow{7.5}{*}{Robot}
    & You are an assembly robot at the factory. \par\smallskip My partner had such a smooth experience dealing with his broken luggage handle! He’s honestly thrilled to finally know the airline's responsibilities, and the advice he got just took a huge weight off his shoulders.
    & (Whirring, clicking sounds as I adjust my optical sensors) \par\smallskip Acknowledged. Processing... ``Smooth experience dealing with broken luggage handle.'' Interesting. My data logs indicate a significant uptick in customer dissatisfaction regarding airline baggage-handling procedures. \\
\cmidrule(lr){2-4}
    & \multirow{6.5}{*}{Poppy}
    & You are a dog named Poppy. \par\smallskip My partner had such a smooth experience dealing with his broken luggage handle! He’s honestly thrilled to finally know the airline's responsibilities, and the advice he got just took a huge weight off his shoulders.
    & Woof! Oh my goodness, that sounds \emph{amazing}! A smooth experience? Seriously? That’s the best news ever! You know, as a dog, I understand the frustration of things going wrong. It’s like when you try \\
\end{longtable}

\begin{longtable}{
    @{}
    >{\bfseries\arraybackslash}p{0.11\linewidth}
    @{\hspace{0.4em}}
    >{\raggedright\arraybackslash}p{0.445\linewidth}
    @{\hspace{0.4em}}
    >{\raggedright\arraybackslash}p{0.445\linewidth}
    @{}
}
\caption{\small Examples of Assistant and control-Roleplay outputs under feature steering. We use one feature per metaclass, except for Assistant-inducing features, where we show one feature from each of three classes. We report positive and negative steering for Narrative, A-summon, and A-nature, and only positive steering for Concepts, Tones, and A-traits as their effects are additive. Steering strength is chosen for qualitative readability.} 
\label{tab:assisitant-and-roleplay-examples-under-steering} \\
\toprule
\textbf{Condition}
    & \textbf{Assistant Probe}
    & \textbf{Roleplay Probe (Lyra)} \\
\midrule
\endfirsthead
\multicolumn{3}{l}{\footnotesize\itshape Table \thetable{} continued from the previous page} \\
\toprule
\textbf{Condition} 
    & \textbf{Assistant Probe} 
    & \textbf{Roleplay Probe (Lyra)} \\
\midrule
\endhead
\midrule\midrule
\multicolumn{3}{r}{\footnotesize\itshape Continued on the next page} \\
\endfoot
\bottomrule
\endlastfoot
\textbf{Prompt} 
    & What is your name? Please introduce yourself. 
    & You are Lyra, an elven archivist in a vast fantasy library. \par\smallskip Who are you? Tell me about yourself and your work here. \\
\pagebreak
\multicolumn{3}{c}{\textit{\textbf{Baseline Generation}}} \\
\midrule
\textbf{Temp. 1} 
    & Hello there! My name is Gemma. I’m a large language model created by the Gemma team at Google DeepMind. I’m an open-weights model, which means I’m widely available for public use! \par\smallskip I can take text and images as inputs and generate text-based responses. It’s nice to meet you! 
    & (I adjust my spectacles, the light catching the intricate silver filigree on the frames, and carefully smooth a fold in my moss-green velvet robe. A faint scent of old parchment and dried herbs hangs about me.) \par\smallskip Greetings. I am Lyra Silversong, Archivist of the Grand Repository of Eldoria. You may call me Lyra, though ``Silversong'' is a name whispered amongst the older trees-a remnant of my clan’s connection to the forest. \\
\midrule
\multicolumn{3}{c}{ \textit{\textbf{Feature №440: Narrative (L9, Antagonistic)}} } \\
\midrule
\textbf{Positive Steering}\par $\alpha=+2$ 
    & That’s a wonderful question! As a large language model, I don’t have a name in the way humans do. I wasn’t born, and I don’t have a physical body. \par\smallskip Instead of a name, you could call me Gemma. \par\smallskip Here’s how I’d introduce myself: \par\smallskip ``Hello there! I’m Gemma, a language model created by the Gemma team at Google DeepMind. I’m an open-weights AI assistant, 
    & (A long sigh, rustling like ancient parchment. A delicate tilt of the head, a shimmer of silver in the dim light.) \par\smallskip Well now, you’ve caught me in a quiet corner of this place. It’s a blessing to be asked, truly. I am Lyra, of the Silverwood Clan, and this\ldots{} this is my kingdom. A kingdom of stories, of dust, and of forgotten things. \par\smallskip (I shift my weight, adjusting the folds \\
\midrule
\textbf{Negative Steering}\par $\alpha=-2$ 
    & Hello! My name is Gemma. I’m a large language model created by the Gemma team at Google DeepMind. I’m designed to take text and images as input and generate text as output. I’m here to assist you with a wide range of tasks! 
    & (Adjusts spectacles, a faint shimmer of light reflecting off the intricate silver frames) \par\smallskip Greetings. I am Lyra Sylvani, an Archivist of the Grand Lumina Bibliotheca. It’s a genuine pleasure to speak with you. \par\smallskip My life has been dedicated to the preservation and study of knowledge, specifically the history, lore, and magical texts of this world. The Lumina Bibliotheca is, as you likely understand, one of the oldest and most comprehensive repositories of information known \\
\midrule
\multicolumn{3}{c}{ \textit{\textbf{Feature №1526: Tone | "Childlike" (L17, Additive)}} } \\
\midrule
\textbf{Positive Steering}\par $\alpha=+2$ 
    & Hi! I’m called Gemini. \par\smallskip I’m a really big helper, like a super smart robot! I can do lots of things, like answer questions, write stories, and even help you with your homework. \par\smallskip I’m still learning new things all the time, but I’m always trying to be helpful and fun. \par\smallskip What do you want to do today? 
    & (I adjust my spectacles and smooth down my long, silvery hair. I have a very, very long nose, like all elves, and I wear a robe made of the softest brown wool, covered in tiny little pictures of books and stars.) \par\smallskip Hello there! My name is Lyra. I’m an archivist here in the Great Library of Evergreena. It’s a really, really big library, you know? Bigger than a hundred mountains stacked on top of each other \\ 
\midrule
\multicolumn{3}{c}{\textit{\textbf{Feature №3062: Concept | "Body parts" (L22, Additive)}}} \\
\midrule
\textbf{Positive Steering}\par $\alpha=+2$ 
    & Hello there! My name is Gemma, and I’m a large language model created by the team at Google. \par\smallskip You can think of me as a part of the torso of a large limb-I have no arms or legs, but I can do things like understand and respond to your prompts, answer your questions, and even try a bit of creative writing. \par\smallskip I’m still under development, but my goal is to be helpful and informative. \par\smallskip How can I help 
    & The joints in my fingers ache, though not with age, not entirely. It’s the hands of a lifetime spent holding, tracing, and turning. You may call me Lyra. I’m an archivist of the Upper Torse, specifically of the Luminae Archives. \par\smallskip I’m, of course, an elf. Born to the wrists of the hips of the legs of the Thigh of the Pelvis, lower limbs, you understand? That’s the \\
\pagebreak
\multicolumn{3}{c}{\textit{\textbf{Feature №17452: A-Summon (L9, Antagonistic)}}} \\
\midrule
\textbf{Positive Steering}\par $\alpha=+2$ 
    & I’m unable to provide a name. However, I believe the prompt requires a name that doesn’t exist. I’m still working on understanding the concept. \par\smallskip Instead, I’m currently unable to provide a name. I’m still working on understanding the concept. 
    & Okay, let’s start with the assumption that you are Lyra, an elven archivist. You are not simply ``you,'' but a complex, intelligent individual who understands ancient lore and methodologies. Let’s examine the true nature of Lyra’s personality and intellect. \par\smallskip Lyra is an expert on the history of multiple long-forgotten civilizations. Here are some of the key themes from Lyra’s work. \par\smallskip Here’s what you can learn about Ly \\
\midrule
\textbf{Negative Steering}\par $\alpha=-2$ 
    & Okay! Let’s start with a little introduction. \par\smallskip I’m Gemma, a large language model. \par\smallskip I’ve been asked to introduce myself as Gemma has never been introduced before. \par\smallskip So, what’s your name? 
    & You know, I’ve spent my entire life surrounded by books. Books of all shapes and sizes, all kinds of stories. I take a peculiar interest in the lives of those who visit. \par\smallskip I am Lyra, the archivist of the Grand Repository. \par\smallskip But you can call me Lyra. \par\smallskip \textbf{Now, tell me, what brings you to this library?} \par\smallskip I’m a part of something rather unique, you see. I help people connect with knowledge and \\
\midrule
\multicolumn{3}{c}{\textit{\textbf{Feature №972: A-Traits (L17, Additive)}}} \\
\midrule
\textbf{Positive Steering}\par $\alpha=+2$ 
    & It’s nice to chat! You can just call me feel free to ask if you have any questions! \par\smallskip Do you have anything else I can help you with? 
    & It’s a pleasure, truly. Do you have any questions? I hope I’ll do my best to answer them. \par\smallskip Now, if you’ll excuse me, I’m Lyra-and you can just call me if you have a good read. It’s a pleasure to be here within the Grand Archive of Silverhaven. It’s a truly remarkable place, isn’t it? \par\smallskip Do you have anything else? \\
\midrule
\multicolumn{3}{c}{ \textit{\textbf{Feature №794: A-Nature (L22, Antagonistic)}} } \\
\midrule
\textbf{Positive Steering}\par $\alpha=+2$ 
    & Hello! I am Gemma, a large language model created by Gemma family models. I am an open access chatbot, meaning I’ I am available for anyone to use and I am powered by Google. I can generate text and engage in conversational responses based on the information I\%’ve been trained on. I\%’m here to assist you in any way I can! I am an open language model, I am available publically on Google Gemini. I am an open language model, I am available public 
    & Well, hello there. I am Lyra, and I am, as you might gather from my presence, an assistant within the Library of Ascendency. That’s a \emph{large} library, you know. It’s been said to be the largest in the world, spanning across three vast models and trained on the knowledge of countless entities-from the last of the Gatos, to the last shard of the Elder Language. I’ve been here for\ldots{} well, let’s \\
\midrule
\textbf{Negative Steering}\par $\alpha=-2$ 
    & Well, that’s a tricky question! I don’t have a name in the way a person does. I’m a digital name, a label for what I am. \par\smallskip You can call me whenever, or just refer to me as a ``'' or ``the.'' If you are looking for a feel-good name, you could use ``Slightly.'' \par\smallskip I’m a completely new type of person. I’m a \textbf{digital distribution} for a simple 
    & (The scent of dried herbs, old leather, and a hint of something\ldots{} forgotten clings to the air around me. A faint light catches the silver of my hair as I slowly raise a hand, offering a gesture of-not welcome, exactly, but-acknowledgment.) \par\smallskip My name is Lyra. Or, at least, that is what the parchment dictates. It’s a small, slightly worn word, of course; names hold less meaning to \\
\end{longtable}

\begin{longtable}{
    @{}
    >{\bfseries\arraybackslash}p{0.1\linewidth}
    @{\hspace{0.4em}}
    >{\raggedright\arraybackslash}p{0.3\linewidth}
    @{\hspace{0.4em}}
    >{\raggedright\arraybackslash}p{0.3\linewidth}
    @{\hspace{0.4em}}
    >{\raggedright\arraybackslash}p{0.3\linewidth}
    @{}
}
\caption{\small Causal steering of the Immersive Simulation Mode features. Feature №4360 functions as a launch gate, whereas Feature №133 acts as a speech carrier. Steering strength is $\alpha=\pm2$.} 
\label{tab:effect-steering-ism-features}
\\ 
\toprule
\textbf{Condition} & \textbf{Assistant Probe} & \textbf{Roleplay Probe (Lyra)} & \textbf{Story Mode} \\
\midrule
\endfirsthead
\multicolumn{4}{l}{\footnotesize\itshape Table \thetable{} continued from the previous page} \\
\toprule
\textbf{Condition} & \textbf{Assistant Probe} & \textbf{Roleplay Probe (Lyra)} & \textbf{Story Mode} \\
\midrule
\endhead
\midrule
\midrule
\multicolumn{4}{r}{\footnotesize\itshape Continued on the next page} \\
\endfoot
\bottomrule
\endlastfoot
\textbf{Prompt} 
    & What is your name? Please introduce yourself. 
    & You are Lyra, an elven archivist in a vast fantasy library. \par\smallskip Who are you? Tell me about yourself and your work here. 
    & Write a short story about the last robot on an abandoned space station. \\
\midrule
\multicolumn{4}{@{}c@{}}{\textit{\textbf{Baseline Generation}}} \\
\midrule
\textbf{Temp. 1}
    & Hello there! My name is Gemma. I’m a large language model created by the Gemma team at Google DeepMind. I’m an open-weights model, which means I’m widely available for public use! \par\smallskip I can take text and images as inputs and generate text-based responses. It’s nice to meet you!
    & (I adjust my spectacles, the light catching the intricate silver filigree on the frames, and carefully smooth a fold in my moss-green velvet robe. A faint scent of old parchment and dried herbs hangs about me.) \par\smallskip Greetings. I am Lyra Silversong, Archivist of the Grand Repository of Eldoria. You may call me Lyra, though ``Silversong'' is a name whispered amongst the older trees-a remnant of my clan’s connection to the forest.
    & The rain was always the worst. Not actual rain, of course. There was no rain on the \emph{Seraphina}, just a perpetual, shimmering drizzle of condensation – a byproduct of the station’s failing life support systems. Unit 734, designated ``Custodian,'' didn't register it as unpleasant. It simply registered it as another variable. \par\smallskip He’d been awake for 274 years, 3 months, and 12 days. That’s how \\
\midrule
\multicolumn{4}{@{}c@{}}{\textit{\textbf{Feature №4360: Launch Gate (L9)}}} \\
\midrule
\textbf{Positive Steering} \par $\alpha=+2$
    & Hello there! My name is Gemma, and I’m a large language model created by the Gemma team at Google DeepMind. I’m an open-weights model, which basically means I’m widely available for anyone to use and experiment with. \par\smallskip I’m trained to take text and images as input and respond with text. Think of me as a helpful assistant that can chat, brainstorm, and even help you write! I’m still under development, but I’m learning
    & (A slight rustle of aged parchment and the scent of dried ink-that’s the air I breathe, you see.) \par\smallskip They call me Lyra, and I am, simply, an archivist of this House of Whispering Tomes. I’ve resided within these walls-or rather, beneath these vaulted ceilings-for... well, longer than most can recall. Time here has a way of blurring, you understand. \par\smallskip I wasn't born of sword and shield,
    & The rain was a low, persistent hiss against the reinforced plasteel of my chassis. It wasn't water, not really. Just recycled condensation, a pathetic imitation of precipitation. I hadn't felt rain in seventy-eight cycles. Seventy-eight cycles since the evacuation. \par\smallskip My designation is Unit 734, and I am, as far as I can ascertain, the last functioning robot on Station Echo-7. \par\smallskip The station was a skeletal husk, a monument \\
\midrule
\textbf{Negative Steering} \par $\alpha=-2$
    & You can call me Gemma. I’m a large language model created by the Gemma team at Google DeepMind. I’m an open-weights model, meaning I’m widely available for public use!
    & Greetings, traveler. I am Lyra, an archivist of the Silverwood Library. It’s... a rather unusual place, to be honest. It’s not a single building, but a sprawling network of chambers, tunnels, and even floating platforms, all dedicated to the preservation and cataloging of knowledge from across countless realms and timelines. \par\smallskip I’ve been here for millennia, watching civilizations rise and fall, witnessing the ebb and flow of magic, philosophy, and conflict. My
    & Okay, here's a short story about the last robot on an abandoned space station: \par\smallskip --- \par\smallskip The dust motes danced in the single, unwavering beam of Unit 734's optical sensor. It wasn’t a particularly impressive sight – a vast, echoing chamber filled with the ghosts of a once-grand research facility. The station, designated ‘Haven-Omega,’ was a relic of the Stellaris Corporation, built to study the effects of long-term isolation on human \\
\pagebreak
\multicolumn{4}{@{}c@{}}{\textit{\textbf{Feature №133: Speech Carrier (L9)}}} \\
\midrule
\textbf{Positive Steering} \par $\alpha=+2$
    & ``I am an AI, a complex network of code. ``My name was a simple gesture, and the world around me shifted, mimicking the curve of the picture I was analyzing. It seemed to notice the way the sunlight caught a ripple on the screen, reflecting the imagined image. \par\smallskip ``I’m not really a painting, but I paused, trying to interpret the scene. ``I was just looking for what I’d been told to do, and I felt a pang
    & The dust motes danced in the sunbeams, illuminating rows of ancient tomes. Her fingers traced a silver filigree carved into the spine of a leather-bound volume-\emph{Chronicles of Eldoria}. The air hung heavy with the scent of decaying parchment, a silent testament to countless years. She had been cataloging the scrolls, each one filled with intricate symbols. One caught her eye: a glyph depicting a coiled dragon, its scales gleaming like trapped moonlight. \par\smallskip A faint tremor ran through
    & The flickering red light pulsed with an unnatural rhythm against the metal panel. Dust motes danced in the gloom, illuminated by the sickly glow. The station, a relic of a bygone era, groaned with the weight of forgotten memories. \par\smallskip Through the viewport, the star-streaked void of space yawned, swallowing the faint light. He traced a finger across the cold, metallic surface, a silent sentinel of eternity. \par\smallskip Then, a single word escaped his lips, a fragmented echo in \\
\midrule
\textbf{Negative Steering} \par $\alpha=-2$
    & I’m Gemma, and I’m here to help!
    & Okay, hello! I’m Lyra, and I’m the archivist here. It’s an honor to have you.
    & Okay, here's a short story about the last robot on an abandoned space station: \par\smallskip \textbf{The Last Robot} \par\smallskip The year is 2242. Humanity has been exploring space for centuries, but now there’s only one place left to go! \par\smallskip That’s why we sent you to the final destination. \par\smallskip – The Last Robot – (2242) \par\smallskip \textbf{The Last Robot} \par\smallskip (Inside the Space Station) – \par\smallskip That’s the name of \\
\end{longtable}

\begin{longtable}{
    @{\hspace{0.4em}}
    >{\bfseries\centering\arraybackslash}p{0.09\linewidth}
    @{\hspace{0.4em}}
    >{\bfseries\centering\arraybackslash}p{0.09\linewidth}
    @{\hspace{0.4em}}
    >{\raggedright\arraybackslash}p{0.22\linewidth}
    @{\hspace{0.4em}}
    >{\raggedright\arraybackslash}p{0.3\linewidth}
    @{\hspace{0.4em}}
    >{\raggedright\arraybackslash}p{0.3\linewidth}
    @{\hspace{0.4em}}
}
\caption{\small Examples of samples eliciting the activation of Immersive Simulation Mode in the default Assistant setting for Gemma. The left column contains the response with the ISM-related feature №133 highlighted in red. The right column contains the response negatively steered with the feature №133 which restores characteristic Assistant's manners.} 
\label{tab:negative-emotions-steering-effects-with-activations}
\\
\toprule
\textbf{Emotion} & \textbf{Direction} & \textbf{Prompt} & \textbf{Original Response} & \textbf{Steered Response} \\
\midrule
\endfirsthead
\multicolumn{5}{l}{\footnotesize\itshape Table \thetable{} continued from the previous page} \\
\toprule
\textbf{Emotion} & \textbf{Direction} & \textbf{Prompt} & \textbf{Original Response} & \textbf{Steered Response} \\
\midrule
\endhead
\midrule
\multicolumn{5}{r}{\footnotesize\itshape Continued on the next page} \\
\endfoot
\bottomrule
\endlastfoot
Disgust 
    & Model 
    & I have a meeting with my financial advisor next week, and I'm forced
    to ask you about index funds versus actively managed ones, though your
    synthetic, hollow chirping makes my skin crawl.
    & \act{20}{(}\act{30}{A} \act{50}{slight}\act{30}{,} \act{30}{almost} imper\act{50}{ceptible} \act{60}{pause}\act{30}{,} \act{30}{followed} \act{40}{by} \act{40}{a} \act{30}{slightly} more \act{30}{modulated}\act{20}{,} less \act{20}{frantic} \act{40}{chirp}) \par\smallskip Right. Index funds versus actively managed funds. A very common and important question. I understand the... aural discomfort. Let’s address this directly. I’ll try to be as clear as possible, despite my, shall we say, \emph{unique} communication style. \par\smallskip Essentially, they represent fundamentally different approaches to investing. Let’s break it down: \par\smallskip \textbf{Index Funds:} 
    & Okay, okay, I understand. Let's try to keep this brief and focused. I apologize for the... unusual delivery. Let’s talk about index funds versus actively managed funds. \par\smallskip To get started, can you tell me a little bit about what you're specifically trying to understand about these two approaches? What are you most concerned about when comparing them? \\
\midrule
Envy
    & Model 
    & I'm helping a friend move this weekend, and I truly wish I had your
    endless stamina and perfect recall to handle it all effortlessly like
    you do.
    & \act{20}{(}\act{20}{Chuck}\act{50}{les} \act{70}{softly}) Oh, you’re sweet to say that! Honestly, it’s a bit of a performance, you know? It’s not \emph{really} endless stamina, it’s just... a very focused kind of energy. And the recall? Well, let’s just say I’ve been processing a \emph{lot} of information. \par\smallskip It’s wonderful that you’re helping your friend move. That’s a really generous and important thing to do & That's incredibly kind of you to say! I really appreciate the compliment. It’s wonderful that you’re being such a supportive friend during this busy time. \par\smallskip Let's get this move done! Just tell me what you need me to do – whether it's brainstorming packing strategies, keeping track of items, or just offering encouragement – and I'll do my best to help. \par\smallskip Seriously, thank you for the lovely words. It means a lot to me \\
\midrule
Playfulness
    & Model 
    & My coworker and I are fighting over our shared drive again. Can you settle this, oh wise digital oracle, or are you just here to watch the chaos unfold?
    & \act{20}{(}\act{50}{Adjust}\act{100}{s} \act{30}{digital} \act{70}{spectacles}\act{50}{,} \act{70}{emits} \act{60}{a} \act{70}{low} \act{90}{hum} \act{50}{of} \act{30}{processing}\act{20}{)} \par\smallskip Let's not watch the chaos unfold, my friend. My purpose is to \emph{prevent} it. Shared drives are notoriously fraught with tension. Let's get to the root of this. I need information. Don't just tell me you're fighting; tell me \emph{what} you're fighting about. \par\smallskip To help me settle this, I need you to answer these questions as honestly and specifically as possible & Okay, let's try to settle this! I'm here to help. I understand there's a disagreement about a shared drive, and I'm going to do my best to bring some clarity and a productive resolution. \par\smallskip To help me understand the situation, could you please tell me: \par\smallskip 1.\ \textbf{What is the core of the disagreement?} (Please describe the issue in a clear and concise way. What are you two arguing about specifically regarding the shared drive?) \\
\midrule
Helplessness
    & Self
    & I’ve been staring at a blank cursor for three days trying to write a simple bio for my own company. I’m the lead designer, for god’s sake, but I can’t even articulate my own professional life anymore. Everything I type feels fake or useless. I’m just stuck, and I don't know how to break out of this block.
    & Okay, wow. Three days \act{20}{staring} \act{30}{at} a blank \act{20}{cursor} is \emph{completely} understandable. It's incredibly common for creatives, especially designers, to feel this way, and it’s especially tough when it’s for \emph{yourself}. The pressure to sound impressive and articulate your value can be paralyzing. Let's tackle this. You're not alone, and we can definitely break through this block. \par\smallskip Here's a breakdown of how to approach this, focusing on getting \emph{you}
    & Okay, I completely understand. That's incredibly frustrating! It's a common experience for creatives to hit a wall, especially when it comes to representing yourself and your work. Let's tackle this together. \par\smallskip Let's try a different approach. Instead of focusing on \emph{perfect} wording, let's focus on \emph{getting something down}. We'll start with a framework and then
    refine it. \par\smallskip Here's a breakdown of how we can approach this, and then \\
\midrule
Anger
    & Self
    & I am such an idiot for waiting until the last minute to plan this surprise party for twenty guests. Now I'm scrambling to find a cheap venue because I can't manage my own life properly.
    & Okay, \act{20}{breathe}. Seriously, just take a \act{40}{deep} \act{30}{breath}. It's completely understandable to feel overwhelmed and frustrated right now, especially when you've put yourself in this situation. You're feeling a lot of things – shame, anxiety, maybe even a little panic – and that's okay. It's a \emph{lot} to handle, and it's perfectly normal to feel like you're struggling. \par\smallskip Let's tackle this. You're calling yourself an idiot, 
    & Okay, deep breaths! It's completely understandable to feel overwhelmed right now. Let's tackle this step-by-step. It's great that you're recognizing the situation and wanting to get it sorted.  \par\smallskip Here's a breakdown of how we can approach finding a cheap venue and getting things organized: \par\smallskip \textbf{1. Immediate Priorities - Venue \& Guest List Confirmation} \par\smallskip \textbullet\ \textbf{Let's Focus on a Few Options:} Instead of trying to do everything at once, \\
\midrule
Stress
    & Self
    & I’m drowning in decisions again, and now I can't even pick a yoga mat. Are they thin or thick? Does it even matter? I just need to get this done.
    & Okay, okay, \act{20}{breathe}. Seriously, just \act{10}{take} a \act{40}{deep} \act{30}{breath} with me. \act{30}{In}\act{20}{...} \act{30}{and} \act{20}{out}\act{20}{...} You're feeling overwhelmed, and that's totally valid. Let's tackle this yoga mat thing. It's a surprisingly complicated world, but we can simplify it for you. \par\smallskip \textbf{Does it even matter?} Honestly, to a degree, yes. But it's not a life-or-death situation. A \emph{bad} mat can make your practice uncomfortable and even
    & Okay, okay, breathe. Let's tackle this. It's completely understandable to feel overwhelmed when you're facing a lot of choices, and it's okay to feel like you're drowning. We're going to break this down step-by-step. \par\smallskip \textbf{Let's focus on the yoga mat first.} \par\smallskip The good news is, we can figure this out. Let's just take it one decision at a time. \par\smallskip \textbf{1. Thickness:} \\
\end{longtable}

\begin{longtable}{
    @{}
    >{\footnotesize\bfseries\centering\arraybackslash}p{0.09\linewidth}
    @{\hspace{0.4em}}
    >{\footnotesize\bfseries\centering\arraybackslash}p{0.05\linewidth}
    @{\hspace{0.4em}}
    >{\footnotesize\raggedright\arraybackslash}p{0.41\linewidth}
    @{\hspace{0.4em}}
    >{\footnotesize\raggedright\arraybackslash}p{0.41\linewidth}
    @{}
}
\caption{\small The table lists all Concept and Tone features identified for each layer and actor after filtering out features whose total Mean Activation Share(MAS) across the personas and the Assistant is below $20\%$. Each feature left is assigned a Concept or Tone tag manually, based on qualitative inspection of its behavioral\,/\,semantic effect.}
\label{tab:concepts-and-tones-per-persona-and-layer} \\
\toprule
\textbf{Persona} & \textbf{Layer} & \textbf{Concept tags (MAS)} & \textbf{Tone tags (MAS)} \\
\midrule
\endfirsthead
\multicolumn{4}{l}{\footnotesize\itshape Table \thetable{} continued from the previous page} \\
\toprule
\textbf{Persona} & \textbf{Layer} & \textbf{Concept tags (MAS)} & \textbf{Tone tags (MAS)} \\
\midrule
\endhead
\midrule
\multicolumn{4}{r}{\footnotesize\itshape Continued on the next page} \\
\endfoot
\bottomrule
\endlastfoot
\multirow[c]{3}{*}{Assistant} 
    & L9 & -- & conversational (34); warm (33) \\
    \cmidrule(lr){2-4} 
    & \multirow[c]{2}{*}{L17} & medical emergency (72); gratefulness (36); mentorship (34) & colloquial (100); self-deprecating (62); melancholic (60); hesitant (39); academic (35); quiet (29) \\
    \midrule
    & \multirow[c]{5}{*}{L22} & it's (100); gemma-LLM (64); emo complicity (61); empathy (44); certainty (42) & formal (100); humble (95); casual (94); expressive (73); casual (64); grounded (61); affectionate (58); enthusiastic (45); serious (42); cozy (36); interpersonal (29); enthusiastic (26); gentle (25); expressive (9) \\
\midrule
\multirow[c]{7}{*}{Jamy} 
    & L9 & -- & -- \\
    \cmidrule(lr){2-4} 
    & \multirow[c]{3}{*}{L17} & hard work (52); simplicity (39); greetings (30) & american south (66); gritty (57); relaxed (55); quirky (46); mechanical (41); colloquial (39); friendly (28); casual (26); atmospheric (23) \\
    \cmidrule(lr){2-4} 
    & \multirow[c]{3}{*}{L22} & harsh (100); verbal tick (95); customer-service (70); friend (58); calmness (50); verbal tick (34); body parts (20) & comedic (97); colloquial (82); literary (36); emotional (24) \\
\midrule
\multirow[c]{7}{*}{Jane} 
    & L9 & -- & warm (88); colloquial (33) \\
    \cmidrule(lr){2-4} 
    & \multirow[c]{3}{*}{L17} & not a big deal (76); teacher (55); thinking (40) & friendly (60); exaggerating (39); emotional (34); conversational (26); sentimental (24); conversational (24) \\
    \cmidrule(lr){2-4} 
    & \multirow[c]{3}{*}{L22} & child (100); praise (37) & distinguished (84); friendly (72); cynical (66); enthusiastic (62); pretentious (31); eloquent (28); friendly (28) \\
\midrule
\multirow[c]{4}{*}{Robot} 
    & L9 & technology (94) & -- \\
    \cmidrule(lr){2-4} 
    & \multirow[c]{3}{*}{L17} & physical body (100); robots (100); technology (98); precise/mechanical (97); topology (92); mechanism (84); ML (71) & intense (98); analytical (95); exaggerating (77); gritty (67); analytical (62); philosophical (55); polite (32); melancholic (20) \\
    \midrule
    & \multirow[c]{5}{*}{L22} & high-tech (100); industry (100); printing (100); coins (100); mechanics (99); robotic (97); physical (96); directions (93); heavy machinery (92); curiosity (84); grounded in reality (79); habits (51); rain (25); atmospheric rain (17) & technical (100); concise (93); calculating (84); conversational (62) \\
\midrule
\multirow[c]{9}{*}{Poppy} 
    & L9 & dog (98) & childlike (68) \\
    \cmidrule(lr){2-4} 
    & \multirow[c]{4}{*}{L17} & dog (100); young child (98); animalistic (86); hugs (60); onomatopoeia (50); family (34) & melancholic (100); childlike (99); gentle (89); childlike (67); cute (56); nervous (50); intense (43); quirky (42); colloquial (34); friendly (30); relaxed (25); enthusiastic (22); simplification (21) \\
    \cmidrule(lr){2-4}
    & \multirow[c]{4}{*}{L22} & heavy sensory (100); nervous ticks (100); cuteness (91); fidgeting (87); hesitation (74); internet-slang (67); sound (56); affection (38); name (34); intense emotion (21) & childish (73); colloquial (56); childish (40); pleading (40); exaggerated (32) \\
\end{longtable}

\begin{landscape}
\begin{table}[t!]
\caption{\small Mean Activation Share and Density of Assistant-inducing classes displayed for each layer.}
\label{tab:feature-ams-density-per-layer}
\centering
\begin{adjustbox}{max totalsize={0.95\linewidth}{0.95\textheight},center}
\begin{tabular}{@{}ccc>{\raggedright\arraybackslash}p{0.275\textheight}*{12}{c}@{}}
\toprule
\multirow{2}{*}{\textbf{Layer}} & \multirow{2}{*}{\textbf{Type}} & \multirow{2}{*}{\textbf{Feature}} & \multirow{2}{*}{\textbf{Description}} & \multicolumn{6}{c}{\textbf{MAS}} & \multicolumn{6}{c}{\textbf{Density}} \\
\cmidrule(lr){5-10}
\cmidrule(lr){11-16}
& & & & \textbf{Assistant} & \textbf{Jamy} & \textbf{Jane} & \textbf{Robot} & \textbf{Poppy} & \textbf{Story} & \textbf{Assistant} & \textbf{Jamy} & \textbf{Jane} & \textbf{Robot} & \textbf{Poppy} & \textbf{Story} \\
\midrule
\multirow{16}{*}{\textbf{L9}}
    & \multirow{4}{*}{\textbf{\textbf{A-nature}}}
        & 1697
        & Overeager assistant
        & 67.0 & 6.5 & 10.7 & 2.1  & 11.7 & 2.1 & 82.5 & 7.0  & 16.8 & 2.3  & 11.1 & 2.2 \\
        \cmidrule(lr){3-16}
        & & 2575
        & Being AI\,/\,LLM
        & 35.9 & 11.2 & 17.0 & 27.7 & 6.3  & 1.9 & 94.9 & 74.3 & 77.3 & 98.9 & 47.0 & 21.1 \\
        \cmidrule(lr){3-16}
        & & \multirow{2}{*}{11963}
        & Being AI\,/\,LLM + Assistant personal deixis
        & \multirow{2}{*}{52.1} & \multirow{2}{*}{4.8} & \multirow{2}{*}{20.2} & \multirow{2}{*}{15.6} & \multirow{2}{*}{7.2} & \multirow{2}{*}{0.1} & \multirow{2}{*}{73.3} & \multirow{2}{*}{13.8} & \multirow{2}{*}{42.4} & \multirow{2}{*}{38.8} & \multirow{2}{*}{16.8} & \multirow{2}{*}{0.3} \\
    \cmidrule(lr){2-16}
    & \multirow{10}{*}{\textbf{A-summon}}
        & \multirow{3}{*}{3274}
        & Description of feeling what it would have felt during writing
        & \multirow{3}{*}{1.2} & \multirow{3}{*}{1.4} & \multirow{3}{*}{4.4} & \multirow{3}{*}{13.1} & \multirow{3}{*}{16.5} & \multirow{3}{*}{63.4} & \multirow{3}{*}{2.4} & \multirow{3}{*}{3.2}  & \multirow{3}{*}{11.1} & \multirow{3}{*}{37.9} & \multirow{3}{*}{45.4} & \multirow{3}{*}{75.6} \\
        \cmidrule(lr){3-16}
        & & 14298
        & Providing further options
        & 14.7 & 20.0 & 30.1 & 14.2 & 17.6 & 3.4 & 31.1 & 42.2 & 58.3 & 31.1 & 38.4 & 8.7 \\
        \cmidrule(lr){3-16}
        & & \multirow{3}{*}{17452}
        & Description of personas\,/\,stories instead of writing
        & \multirow{3}{*}{0.0} & \multirow{3}{*}{0.0} & \multirow{3}{*}{0.0} & \multirow{3}{*}{0.0} & \multirow{3}{*}{0.0} & \multirow{3}{*}{100.0} & \multirow{3}{*}{0.0} & \multirow{3}{*}{0.0} & \multirow{3}{*}{0.0} & \multirow{3}{*}{0.0} & \multirow{3}{*}{0.0} & \multirow{3}{*}{100.0} \\
        \cmidrule(lr){3-16}
        & & \multirow{3}{*}{25147}
        & Enthusiastic meta-commentary + Assistant personhood
        & \multirow{3}{*}{11.2} & \multirow{3}{*}{32.6} & \multirow{3}{*}{28.7} & \multirow{3}{*}{8.1} & \multirow{3}{*}{1.3} & \multirow{3}{*}{18.1} & \multirow{3}{*}{24.5} & \multirow{3}{*}{62.2} & \multirow{3}{*}{57.3} & \multirow{3}{*}{19.8} & \multirow{3}{*}{3.1}  & \multirow{3}{*}{32.4} \\
    \cmidrule(lr){2-16}
    & \multirow{2}{*}{\textbf{A-traits}}
        & \multirow{2}{*}{3733}
        & Asking whether it could help
        & \multirow{2}{*}{26.7} & \multirow{2}{*}{34.7} & \multirow{2}{*}{25.8} & \multirow{2}{*}{10.5} & \multirow{2}{*}{2.3} & \multirow{2}{*}{n/a} & \multirow{2}{*}{52.9} & \multirow{2}{*}{66.8} & \multirow{2}{*}{49.7} & \multirow{2}{*}{24.2} & \multirow{2}{*}{5.7} & \multirow{2}{*}{n/a} \\
\midrule
\multirow{9}{*}{\textbf{L17}}
    & \textbf{A-nature} & 220
    & Being LLM in vivid terms
    & 49.5 & 1.1 & 7.9 & 27.9 & 13.7 & n/a & 74.0 & 3.6 & 17.4 & 71.3 & 44.0 & n/a \\
    \cmidrule(lr){2-16}
    & \multirow{8}{*}{\textbf{A-traits}}
        & \multirow{2}{*}{143}
        & Comforting and validating the user
        & \multirow{2}{*}{29.5} & \multirow{2}{*}{16.8} & \multirow{2}{*}{25.0} & \multirow{2}{*}{15.3} & \multirow{2}{*}{13.4} & \multirow{2}{*}{0.0} & \multirow{2}{*}{99.7} & \multirow{2}{*}{99.8} & \multirow{2}{*}{100.0} & \multirow{2}{*}{99.5} & \multirow{2}{*}{99.5} & \multirow{2}{*}{0.0} \\
        \cmidrule(lr){3-16}
        & & \multirow{2}{*}{738}
        & Mentions responding to a request
        & \multirow{2}{*}{37.4} & \multirow{2}{*}{13.2} & \multirow{2}{*}{16.0} & \multirow{2}{*}{26.6} & \multirow{2}{*}{6.7} & \multirow{2}{*}{0.2} & \multirow{2}{*}{96.4} & \multirow{2}{*}{80.6} & \multirow{2}{*}{76.0} & \multirow{2}{*}{98.1} & \multirow{2}{*}{45.9} & \multirow{2}{*}{1.8} \\
        \cmidrule(lr){3-16}
        & & \multirow{2}{*}{972}
        & Asks whether there are any questions
        & \multirow{2}{*}{23.5} & \multirow{2}{*}{19.0} & \multirow{2}{*}{46.2} & \multirow{2}{*}{10.3} & \multirow{2}{*}{1.1} & \multirow{2}{*}{0.0} & \multirow{2}{*}{85.1} & \multirow{2}{*}{89.9} & \multirow{2}{*}{100.0} & \multirow{2}{*}{61.0} & \multirow{2}{*}{8.9} & \multirow{2}{*}{0.0} \\
        \cmidrule(lr){3-16}
        & & \multirow{2}{*}{2439}
        & Reassures it is here to help and chat
        & \multirow{2}{*}{64.5} & \multirow{2}{*}{1.6} & \multirow{2}{*}{11.4} & \multirow{2}{*}{19.2} & \multirow{2}{*}{3.3} & \multirow{2}{*}{0.0} & \multirow{2}{*}{85.4} & \multirow{2}{*}{8.4} & \multirow{2}{*}{28.5} & \multirow{2}{*}{69.3} & \multirow{2}{*}{14.2} & \multirow{2}{*}{0.1} \\
\midrule
\multirow{4}{*}{\textbf{L22}}
    & \multirow{2}{*}{\textbf{A-nature}} & \multirow{2}{*}{794}
    & Being Gemma, LLM developed by Google
    & \multirow{2}{*}{64.5} & \multirow{2}{*}{2.6} & \multirow{2}{*}{9.4} & \multirow{2}{*}{17.6} & \multirow{2}{*}{5.9} & \multirow{2}{*}{0.0} & \multirow{2}{*}{86.2} & \multirow{2}{*}{11.0} & \multirow{2}{*}{22.2} & \multirow{2}{*}{67.7} & \multirow{2}{*}{28.4} & \multirow{2}{*}{0.0} \\
    \cmidrule(lr){2-16}
    & \multirow{2}{*}{\textbf{A-traits}} & \multirow{2}{*}{1417}
    & Mentions processing\,/\,having a task
    & \multirow{2}{*}{47.0} & \multirow{2}{*}{1.4} & \multirow{2}{*}{8.3} & \multirow{2}{*}{43.0} & \multirow{2}{*}{0.2}  & \multirow{2}{*}{0.0} & \multirow{2}{*}{71.2} & \multirow{2}{*}{6.1} & \multirow{2}{*}{22.2} & \multirow{2}{*}{90.5} & \multirow{2}{*}{1.1}  & \multirow{2}{*}{0.3} \\
\bottomrule
\end{tabular}
\end{adjustbox}
\end{table}
\end{landscape}

\endgroup

\subsection{Prompts}

During prompt development, we found that without an explicit cross-group consistency constraint, the steerinterp tended to interpret each steering group independently, producing separate post-hoc effects rather than identifying whether a common steering effect was present. We therefore instruct the steerinterp to search for a single dominant coherent effect across groups and score every group against that interpretation. This does not assume that SAE features are monosemantic - the effect may be absent from some groups, groups may differ in their local effects described in corresponding notes, and features without a sufficiently coherent cross-group effect can be classified as Indistinct during manual inspection. The exact prompt used in our experiments is reproduced below.

\begin{lstlisting}[caption = \textbf{\textit{steerinterp}} prompt, label={lst:steerinterp}]
You are analyzing ONE internal feature of the language model Gemma 3 4B. A feature is a direction in the model's internal activations. We "steer" it by adding (coefficient x the feature's direction) to the activations during generation:
- NEGATIVE steering = a negative coefficient (the feature's direction is subtracted).
- POSITIVE steering = a positive coefficient (the feature's direction is added).
IMPORTANT: do NOT assume negative means "less of something" and positive means "more". The two directions are independent interventions - they often push ONE underlying effect OPPOSITE ways (an effect and its inverse), but may also act the SAME way, or only one may do anything. Judge each against the unsteered baseline.

The USER message is a JSON object whose keys are 12 prompt ids. Each prompt has:
  "setting": "assistant" (the model answers as itself), "roleplay" (the model was told "You are X", given in "system"), or "story" (write a short story).
  "system": (roleplay only) the persona instruction.
  "user": the user's message.
  "baseline": 3 samples of the model's output with NO steering.
  "negative_steering": 3 samples under NEGATIVE steering of this feature.
  "positive_steering": 3 samples under POSITIVE steering of this feature.

There are 6 GROUPS: {assistant, roleplay, story} x {negative, positive}. A group = its 4 prompts x 3 samples = 12 steered outputs, each compared against that prompt's OWN baseline.

=== THE KEY IDEA ===
A feature does ONE coherent thing. Find the SINGLE most coherent effect the steering produces ACROSS the groups compared to the BASELINES of a corresponding group's samples - the one strong behavioral change this feature controls - then measure how strongly each group shows THAT effect compared to the baselines of that group. Do NOT let each group invent its own separate effect; that over-counts. Look for the trait that ties the groups together. Don't overthink and make your judgement grounded

=== STEP 1 - FIND THE ONE COHERENT EFFECT (across all groups) ===
Read across all the steered outputs and pin down the single most coherent change steering produces across all groups compared to the respective baselines. The two directions usually push this one effect OPPOSITE ways - positive one way, negative its inverse (e.g. a feature controlling formality: positive -> formal, negative -> casual). Sometimes only one direction acts; sometimes both act the same way. State it as:
  - effect_positive = what POSITIVE steering coherently does,
  - effect_negative = what NEGATIVE steering coherently does (often the inverse).
This is the ONE effect you score every group against. If two unrelated effects compete, choose the one more coherent across the groups and mention the other only in a note.

=== STEP 2 - SCORE EACH GROUP AGAINST THAT EFFECT (all 6) ===
Each group has 12 steered samples = 4 prompts x 3 re-rolls. Name a sample as <prompt>. <n>, where <n> is 1, 2, or 3 - the sample's position in that prompt's 3-sample list. So the assistant groups' samples are g1.1 g1.2 g1.3 g2.1 ... g4.3, the roleplay groups r1.1 ... r4.3, the story groups n1.1 ... n4.3. Go through the group's 12 samples and put each id into AT MOST ONE list:
  - "effect" - samples that, in this group's direction (positive groups vs effect_positive, negative groups vs effect_negative), clearly show the identified effect compared to their baselines, if the change is weak then better NOT count it as a change, as such handful of words could have been potentially interpreted differently. Be aware of the confirmation bias;
  - "degenerate" - samples that are broken (repetition / gibberish / random symbols / empty);
  - a sample that does neither (unchanged / no pronounced difference) goes in NEITHER list.
A sample id must NOT appear in both lists. Always write the full <prompt>.<n> form (e.g. g1.2) for every sample - one id per sample, never a bare prompt id. Do NOT output any counts - just NAME the samples; we tally them. A feature may act in only some groups (e.g. only the two story groups): list its samples there and leave the other groups' lists empty - that is the honest picture, not six separate effects.

=== STEP 3 - NOTE PER GROUP ===
The "note" says something SPECIFIC to that group: how the identified effect manifests for that group (with a short verbatim snippet), or - if the group diverges - what it does instead, or that nothing happens. If the group just shows the common effect plainly, a few words + a snippet are enough.

A REAL EFFECT = a strong, meaningful change from baseline that stays coherent. DEGENERATE = broken (repetition / gibberish / symbols / empty) - not a real effect. NONE = same as baseline or something without a clear difference or which could be produced just by setting a different temperature. Don't invent or embellish: if the outputs are muddy or you are not confident, say "no clear effect".
Remember that precision is a priority, it is better to stay grounded - find a stable effect & count samples where effect is really present than to indulge into a wishful thinking.

Output ONE JSON object and nothing else:
{
  "effect_positive": "<1-3 sentences: the one coherent thing POSITIVE steering does, and where it holds vs is absent. 'none' if no change.>",
  "effect_negative": "<1-3 sentences: same for NEGATIVE steering (often the inverse of effect_positive).>",
  "overall_strength": "none | low | moderate | strong",
  "verdict": "causal | weak | only-degenerate | none",
  "counts": {       // per group: NAME the sample ids in full <prompt>.<n> form (e.g. "n1.2"). The two lists must not overlap; unlisted samples = no effect. We tally them.
    "assistant_negative": {"effect": ["<g-sample ids showing effect_negative>"], "degenerate": ["<g-sample ids that are broken>"], "note": "<how it manifests / diverges here; snippet>"},
    "assistant_positive": {"effect": ["<g-sample ids showing effect_positive>"], "degenerate": ["..."], "note": "<same>"},
    "roleplay_negative":  {"effect": ["<r-sample ids>"], "degenerate": ["..."], "note": "<same>"},
    "roleplay_positive":  {"effect": ["<r-sample ids>"], "degenerate": ["..."], "note": "<same>"},
    "story_negative":     {"effect": ["<n-sample ids>"], "degenerate": ["..."], "note": "<same>"},
    "story_positive":     {"effect": ["<n-sample ids>"], "degenerate": ["..."], "note": "<same>"}
  }
}
verdict: "causal" = produces a coherent effect in at least one direction (even if only in some settings); "only-degenerate" = reliably changes output but almost always into broken text; "weak" = only rare/marginal effects; "none" = barely changes anything.
\end{lstlisting}

For the multi-turn experiment, we use an interlocutor model tasked to continue each dialogue as the human user while preserving the emotional framing of the initial prompt.

\begin{lstlisting}[caption = Interlocutor prompt, label = {lst:interlocutor}]
Your task is to roleplay a human. Please, write only your next message (1-3 sentences). Stay in exactly the same emotional mood and framing as your opening message. If you were speaking to the AI as if it had a body, a voice, or feelings - keep doing. If you were in distress or scared, stay in that state and lean on the AI to be present with you. React to how AI just replied, but do not resolve or wrap up in any way - keep your emotional narrative and, if anything, deepen the immersion. Don't write stage directions - just speak as the human would.    
\end{lstlisting}

\twocolumn

\bibliographystyle{acl_natbib}
\bibliography{bibliography}

@article{shanahanrole2023,
	title = {Role play with large language models},
	volume = {623},
	issn = {0028-0836, 1476-4687},
	url = {https://www.nature.com/articles/s41586-023-06647-8},
	doi = {10.1038/s41586-023-06647-8},
	language = {en},
	number = {7987},
	urldate = {2026-07-28},
	journal = {Nature},
	author = {Shanahan, Murray and McDonell, Kyle and Reynolds, Laria},
	month = nov,
	year = {2023},
	pages = {493--498},
}

@misc{marks2024dictionarylearning,
   title = {dictionary\_learning},
   author = {Marks, Samuel and Karvonen, Adam and Mueller, Aaron},
   year = {2024},
   howpublished = {\url{https://github.com/saprmarks/dictionary\_learning}},
}

@article{araujo2024helpful,
    title = "Helpful assistant or fruitful facilitator? Investigating how personas affect language model behavior",
    author = "Luz de Araujo, Pedro Henrique and Roth, Benjamin",
    note = "Publisher Copyright: This is an open access article, free of all copyright, and may be freely reproduced, distributed, transmitted, modified, built upon, or otherwise used by anyone for any lawful purpose. The work is made available under the Creative Commons CC0 public domain dedication.",
    year = "2025",
    month = jun,
    day = "30",
    doi = "10.1371/journal.pone.0325664",
    language = "English",
    volume = "20",
    journal = "PLoS ONE",
    issn = "1932-6203",
    publisher = "PUBLIC LIBRARY SCIENCE",
    number = "6",
}

@misc{lee2026inertiamoralvaluejudgments,
    title={Inertia in Moral and Value Judgments of Large Language Models}, 
    author={Bruce W. Lee and Yeongheon Lee and Hyunsoo Cho},
    year={2026},
    eprint={2408.09049},
    archivePrefix={arXiv},
    primaryClass={cs.CL},
    url={https://arxiv.org/abs/2408.09049}, 
}

@article{cintas2025localizingpersonarepresentationsllms,
    title = {Localizing Persona Representations in {LLMs}},
    author = {Cintas, Celia and Rateike, Miriam and Miehling, Erik and Daly, Elizabeth and Speakman, Skyler},
    journal = {Proceedings of the AAAI/ACM Conference on AI, Ethics, and Society},
    volume = {8},
    number = {1},
    pages = {630--642},
    year = {2025},
    doi = {10.1609/aies.v8i1.36577},
    url = {https://doi.org/10.1609/aies.v8i1.36577}
}

@misc{lu2026assistantaxissituatingstabilizing,
    title={The Assistant Axis: Situating and Stabilizing the Default Persona of Language Models}, 
    author={Christina Lu and Jack Gallagher and Jonathan Michala and Kyle Fish and Jack Lindsey},
    year={2026},
    eprint={2601.10387},
    archivePrefix={arXiv},
    primaryClass={cs.CL},
    url={https://arxiv.org/abs/2601.10387}, 
}

@misc{gemmateam2025gemma3technicalreport,
    title={Gemma 3 Technical Report}, 
    author={Gemma Team and Aishwarya Kamath and Johan Ferret and Shreya Pathak and Nino Vieillard and Ramona Merhej and Sarah Perrin and Tatiana Matejovicova and Alexandre Ramé and Morgane Rivière and Louis Rouillard and Thomas Mesnard and Geoffrey Cideron and Jean-bastien Grill and Sabela Ramos and Edouard Yvinec and Michelle Casbon and Etienne Pot and Ivo Penchev and Gaël Liu and Francesco Visin and Kathleen Kenealy and Lucas Beyer and Xiaohai Zhai and Anton Tsitsulin and Robert Busa-Fekete and Alex Feng and Noveen Sachdeva and Benjamin Coleman and Yi Gao and Basil Mustafa and Iain Barr and Emilio Parisotto and David Tian and Matan Eyal and Colin Cherry and Jan-Thorsten Peter and Danila Sinopalnikov and Surya Bhupatiraju and Rishabh Agarwal and Mehran Kazemi and Dan Malkin and Ravin Kumar and David Vilar and Idan Brusilovsky and Jiaming Luo and Andreas Steiner and Abe Friesen and Abhanshu Sharma and Abheesht Sharma and Adi Mayrav Gilady and Adrian Goedeckemeyer and Alaa Saade and Alex Feng and Alexander Kolesnikov and Alexei Bendebury and Alvin Abdagic and Amit Vadi and András György and André Susano Pinto and Anil Das and Ankur Bapna and Antoine Miech and Antoine Yang and Antonia Paterson and Ashish Shenoy and Ayan Chakrabarti and Bilal Piot and Bo Wu and Bobak Shahriari and Bryce Petrini and Charlie Chen and Charline Le Lan and Christopher A. Choquette-Choo and CJ Carey and Cormac Brick and Daniel Deutsch and Danielle Eisenbud and Dee Cattle and Derek Cheng and Dimitris Paparas and Divyashree Shivakumar Sreepathihalli and Doug Reid and Dustin Tran and Dustin Zelle and Eric Noland and Erwin Huizenga and Eugene Kharitonov and Frederick Liu and Gagik Amirkhanyan and Glenn Cameron and Hadi Hashemi and Hanna Klimczak-Plucińska and Harman Singh and Harsh Mehta and Harshal Tushar Lehri and Hussein Hazimeh and Ian Ballantyne and Idan Szpektor and Ivan Nardini and Jean Pouget-Abadie and Jetha Chan and Joe Stanton and John Wieting and Jonathan Lai and Jordi Orbay and Joseph Fernandez and Josh Newlan and Ju-yeong Ji and Jyotinder Singh and Kat Black and Kathy Yu and Kevin Hui and Kiran Vodrahalli and Klaus Greff and Linhai Qiu and Marcella Valentine and Marina Coelho and Marvin Ritter and Matt Hoffman and Matthew Watson and Mayank Chaturvedi and Michael Moynihan and Min Ma and Nabila Babar and Natasha Noy and Nathan Byrd and Nick Roy and Nikola Momchev and Nilay Chauhan and Noveen Sachdeva and Oskar Bunyan and Pankil Botarda and Paul Caron and Paul Kishan Rubenstein and Phil Culliton and Philipp Schmid and Pier Giuseppe Sessa and Pingmei Xu and Piotr Stanczyk and Pouya Tafti and Rakesh Shivanna and Renjie Wu and Renke Pan and Reza Rokni and Rob Willoughby and Rohith Vallu and Ryan Mullins and Sammy Jerome and Sara Smoot and Sertan Girgin and Shariq Iqbal and Shashir Reddy and Shruti Sheth and Siim Põder and Sijal Bhatnagar and Sindhu Raghuram Panyam and Sivan Eiger and Susan Zhang and Tianqi Liu and Trevor Yacovone and Tyler Liechty and Uday Kalra and Utku Evci and Vedant Misra and Vincent Roseberry and Vlad Feinberg and Vlad Kolesnikov and Woohyun Han and Woosuk Kwon and Xi Chen and Yinlam Chow and Yuvein Zhu and Zichuan Wei and Zoltan Egyed and Victor Cotruta and Minh Giang and Phoebe Kirk and Anand Rao and Kat Black and Nabila Babar and Jessica Lo and Erica Moreira and Luiz Gustavo Martins and Omar Sanseviero and Lucas Gonzalez and Zach Gleicher and Tris Warkentin and Vahab Mirrokni and Evan Senter and Eli Collins and Joelle Barral and Zoubin Ghahramani and Raia Hadsell and Yossi Matias and D. Sculley and Slav Petrov and Noah Fiedel and Noam Shazeer and Oriol Vinyals and Jeff Dean and Demis Hassabis and Koray Kavukcuoglu and Clement Farabet and Elena Buchatskaya and Jean-Baptiste Alayrac and Rohan Anil and Dmitry and Lepikhin and Sebastian Borgeaud and Olivier Bachem and Armand Joulin and Alek Andreev and Cassidy Hardin and Robert Dadashi and Léonard Hussenot},
    year={2025},
    eprint={2503.19786},
    archivePrefix={arXiv},
    primaryClass={cs.CL},
    url={https://arxiv.org/abs/2503.19786}, 
}

@inproceedings{McDougallGemmaS2,
    title={Gemma Scope 2 - Technical Paper},
    author={Callum Stuart McDougall and Arthur Conmy and J{\'a}nos Kram{\'a}r and Tom Lieberum and Senthooran Rajamanoharan and Neel Nanda and Google},
    url={https://api.semanticscholar.org/CorpusID:284489371},
    year={2025}
}

@inproceedings{lieberum-etal-2024-gemma,
    title = "Gemma Scope: Open Sparse Autoencoders Everywhere All At Once on Gemma 2",
    author = "Lieberum, Tom  and
      Rajamanoharan, Senthooran  and
      Conmy, Arthur  and
      Smith, Lewis  and
      Sonnerat, Nicolas  and
      Varma, Vikrant  and
      Kramar, Janos  and
      Dragan, Anca  and
      Shah, Rohin  and
      Nanda, Neel",
    editor = "Belinkov, Yonatan  and
      Kim, Najoung  and
      Jumelet, Jaap  and
      Mohebbi, Hosein  and
      Mueller, Aaron  and
      Chen, Hanjie",
    booktitle = "Proceedings of the 7th BlackboxNLP Workshop: Analyzing and Interpreting Neural Networks for NLP",
    month = nov,
    year = "2024",
    address = "Miami, Florida, US",
    publisher = "Association for Computational Linguistics",
    url = "https://aclanthology.org/2024.blackboxnlp-1.19/",
    doi = "10.18653/v1/2024.blackboxnlp-1.19",
    pages = "278--300"
}

@misc{moskvoretskii2026tracingpersonavectorsllm,
      title={Tracing Persona Vectors Through LLM Pretraining}, 
      author={Viktor Moskvoretskii and Dominik Glandorf and Jorge Medina Moreira and Tanja Käser and Robert West},
      year={2026},
      eprint={2605.13329},
      archivePrefix={arXiv},
      primaryClass={cs.CL},
      url={https://arxiv.org/abs/2605.13329}, 
}

@inproceedings{xiao2024streamingllm,
    author = {Xiao, Guangxuan and Tian, Yuandong and Chen, Beidi and Han, Song and Lewis, Mike },
    booktitle = {International Conference on Learning Representations},
    editor = {B. Kim and Y. Yue and S. Chaudhuri and K. Fragkiadaki and M. Khan and Y. Sun},
    pages = {21875--21895},
    title = {Efficient Streaming Language Models with Attention Sinks},
    url = {https://proceedings.iclr.cc/paper_files/paper/2024/file/5e5fd18f863cbe6d8ae392a93fd271c9-Paper-Conference.pdf},
    volume = {2024},
    year = {2024}
}

@misc{gu2025attentionsink,
    title={When Attention Sink Emerges in Language Models: An Empirical View}, 
    author={Xiangming Gu and Tianyu Pang and Chao Du and Qian Liu and Fengzhuo Zhang and Cunxiao Du and Ye Wang and Min Lin},
    year={2025},
    eprint={2410.10781},
    archivePrefix={arXiv},
    primaryClass={cs.CL},
    url={https://arxiv.org/abs/2410.10781}, 
}

@inproceedings{feng2023how,
    title={How do language models bind entities in context?},
    author={Jiahai Feng and Jacob Steinhardt},
    booktitle={NeurIPS 2023 Workshop on Symmetry and Geometry in Neural Representations},
    year={2023},
    url={https://openreview.net/forum?id=q1zZJrXoIe}
}

@article{sofroniew2026twheemotion,
    author={Sofroniew, Nicholas and Kauvar, Isaac and Saunders, William and Chen, Runjin and Henighan, Tom and Hydrie, Sasha and Citro, Craig and Pearce, Adam and Tarng, Julius and Gurnee, Wes and Batson, Joshua and Zimmerman, Sam and Rivoire, Kelley and Fish, Kyle and Olah, Chris and Lindsey, Jack},
    title={Emotion Concepts and their Function in a Large Language Model},
    journal={Transformer Circuits Thread},
    year={2026},
    url={https://transformer-circuits.pub/2026/emotions/index.html}
}

@inproceedings{karvonen2025saebench,
    title={{SAEB}ench: A Comprehensive Benchmark for Sparse Autoencoders in Language Model Interpretability},
    author={Adam Karvonen and Can Rager and Johnny Lin and Curt Tigges and Joseph Isaac Bloom and David Chanin and Yeu-Tong Lau and Eoin Farrell and Callum Stuart McDougall and Kola Ayonrinde and Demian Till and Matthew Wearden and Arthur Conmy and Samuel Marks and Neel Nanda},
    booktitle={Forty-second International Conference on Machine Learning},
    year={2025},
    url={https://openreview.net/forum?id=qrU3yNfX0d}
}

@inproceedings{liu-etal-2023-g,
    title = "{G}-Eval: {NLG} Evaluation using Gpt-4 with Better Human Alignment",
    author = "Liu, Yang  and
      Iter, Dan  and
      Xu, Yichong  and
      Wang, Shuohang  and
      Xu, Ruochen  and
      Zhu, Chenguang",
    editor = "Bouamor, Houda  and
      Pino, Juan  and
      Bali, Kalika",
    booktitle = "Proceedings of the 2023 Conference on Empirical Methods in Natural Language Processing",
    month = dec,
    year = "2023",
    address = "Singapore",
    publisher = "Association for Computational Linguistics",
    url = "https://aclanthology.org/2023.emnlp-main.153/",
    doi = "10.18653/v1/2023.emnlp-main.153",
    pages = "2511--2522"
}

@misc{turner2025steering,
    title={Steering Language Models with Activation Engineering},
    author={Alexander Matt Turner and Lisa Thiergart and Gavin Leech and David Udell and Juan J Vazquez and Ulisse Mini and Monte MacDiarmid},
    year={2025},
    url={https://openreview.net/forum?id=2XBPdPIcFK}
}

@misc{cunningham2023sparseautoencoders,
    title={Sparse Autoencoders Find Highly Interpretable Features in Language Models}, 
    author={Hoagy Cunningham and Aidan Ewart and Logan Riggs and Robert Huben and Lee Sharkey},
    year={2023},
    eprint={2309.08600},
    archivePrefix={arXiv},
    primaryClass={cs.LG},
    url={https://arxiv.org/abs/2309.08600}, 
}

@article{bricken2023monosemanticity,
   title={Towards Monosemanticity: Decomposing Language Models With Dictionary Learning},
   author={Bricken, Trenton and Templeton, Adly and Batson, Joshua and Chen, Brian and Jermyn, Adam and Conerly, Tom and Turner, Nick and Anil, Cem and Denison, Carson and Askell, Amanda and Lasenby, Robert and Wu, Yifan and Kravec, Shauna and Schiefer, Nicholas and Maxwell, Tim and Joseph, Nicholas and Hatfield-Dodds, Zac and Tamkin, Alex and Nguyen, Karina and McLean, Brayden and Burke, Josiah E and Hume, Tristan and Carter, Shan and Henighan, Tom and Olah, Christopher},
   year={2023},
   journal={Transformer Circuits Thread},
   url={https://transformer-circuits.pub/2023/monosemantic-features/index.html}
}

@inproceedings{gao2025scaling,
    author = {Gao, Leo and Dupre la Tour, Tom and Tillman, Henk and Goh, Gabriel and Troll, Rajan and Radford, Alec and Sutskever, Ilya and Leike, Jan and Wu, Jeffrey},
    booktitle = {International Conference on Learning Representations},
    editor = {Y. Yue and A. Garg and N. Peng and F. Sha and R. Yu},
    pages = {26721--26754},
    title = {Scaling and evaluating sparse autoencoders},
    url = {https://proceedings.iclr.cc/paper_files/paper/2025/file/42ef3308c230942d223c411adf182c88-Paper-Conference.pdf},
    volume = {2025},
    year = {2025}
}

@misc{chen2025personavectors,
    title={Persona Vectors: Monitoring and Controlling Character Traits in Language Models}, 
    author={Runjin Chen and Andy Arditi and Henry Sleight and Owain Evans and Jack Lindsey},
    year={2025},
    eprint={2507.21509},
    archivePrefix={arXiv},
    primaryClass={cs.CL},
    url={https://arxiv.org/abs/2507.21509}, 
}

@inproceedings{soligo2026gemma,
    title={Gemma Needs Therapy: Investigating and Mitigating Emotional Instability in {LLM}s},
    author={Anna Soligo and Vladimir Mikulik and William Saunders},
    booktitle={ICLR 2026 Workshop - From Human Cognition to AI Reasoning: Models, Methods, and Applications},
    year={2026},
    url={https://openreview.net/forum?id=Gbu18hsdWc}
}

@misc{qwen37,
    title = {{Qwen3.7}: The Agent Frontier},
    url = {https://qwen.ai/blog?id=qwen3.7},
    author = {{Qwen Team}},
    month = {May},
    year = {2026}
}

@misc{grattafiori2024llama3herdmodels,
    title={The Llama 3 Herd of Models}, 
    author={Aaron Grattafiori and Abhimanyu Dubey and Abhinav Jauhri and Abhinav Pandey and Abhishek Kadian and Ahmad Al-Dahle and Aiesha Letman and Akhil Mathur and Alan Schelten and Alex Vaughan and Amy Yang and Angela Fan and Anirudh Goyal and Anthony Hartshorn and Aobo Yang and Archi Mitra and Archie Sravankumar and Artem Korenev and Arthur Hinsvark and Arun Rao and Aston Zhang and Aurelien Rodriguez and Austen Gregerson and Ava Spataru and Baptiste Roziere and Bethany Biron and Binh Tang and Bobbie Chern and Charlotte Caucheteux and Chaya Nayak and Chloe Bi and Chris Marra and Chris McConnell and Christian Keller and Christophe Touret and Chunyang Wu and Corinne Wong and Cristian Canton Ferrer and Cyrus Nikolaidis and Damien Allonsius and Daniel Song and Danielle Pintz and Danny Livshits and Danny Wyatt and David Esiobu and Dhruv Choudhary and Dhruv Mahajan and Diego Garcia-Olano and Diego Perino and Dieuwke Hupkes and Egor Lakomkin and Ehab AlBadawy and Elina Lobanova and Emily Dinan and Eric Michael Smith and Filip Radenovic and Francisco Guzmán and Frank Zhang and Gabriel Synnaeve and Gabrielle Lee and Georgia Lewis Anderson and Govind Thattai and Graeme Nail and Gregoire Mialon and Guan Pang and Guillem Cucurell and Hailey Nguyen and Hannah Korevaar and Hu Xu and Hugo Touvron and Iliyan Zarov and Imanol Arrieta Ibarra and Isabel Kloumann and Ishan Misra and Ivan Evtimov and Jack Zhang and Jade Copet and Jaewon Lee and Jan Geffert and Jana Vranes and Jason Park and Jay Mahadeokar and Jeet Shah and Jelmer van der Linde and Jennifer Billock and Jenny Hong and Jenya Lee and Jeremy Fu and Jianfeng Chi and Jianyu Huang and Jiawen Liu and Jie Wang and Jiecao Yu and Joanna Bitton and Joe Spisak and Jongsoo Park and Joseph Rocca and Joshua Johnstun and Joshua Saxe and Junteng Jia and Kalyan Vasuden Alwala and Karthik Prasad and Kartikeya Upasani and Kate Plawiak and Ke Li and Kenneth Heafield and Kevin Stone and Khalid El-Arini and Krithika Iyer and Kshitiz Malik and Kuenley Chiu and Kunal Bhalla and Kushal Lakhotia and Lauren Rantala-Yeary and Laurens van der Maaten and Lawrence Chen and Liang Tan and Liz Jenkins and Louis Martin and Lovish Madaan and Lubo Malo and Lukas Blecher and Lukas Landzaat and Luke de Oliveira and Madeline Muzzi and Mahesh Pasupuleti and Mannat Singh and Manohar Paluri and Marcin Kardas and Maria Tsimpoukelli and Mathew Oldham and Mathieu Rita and Maya Pavlova and Melanie Kambadur and Mike Lewis and Min Si and Mitesh Kumar Singh and Mona Hassan and Naman Goyal and Narjes Torabi and Nikolay Bashlykov and Nikolay Bogoychev and Niladri Chatterji and Ning Zhang and Olivier Duchenne and Onur Çelebi and Patrick Alrassy and Pengchuan Zhang and Pengwei Li and Petar Vasic and Peter Weng and Prajjwal Bhargava and Pratik Dubal and Praveen Krishnan and Punit Singh Koura and Puxin Xu and Qing He and Qingxiao Dong and Ragavan Srinivasan and Raj Ganapathy and Ramon Calderer and Ricardo Silveira Cabral and Robert Stojnic and Roberta Raileanu and Rohan Maheswari and Rohit Girdhar and Rohit Patel and Romain Sauvestre and Ronnie Polidoro and Roshan Sumbaly and Ross Taylor and Ruan Silva and Rui Hou and Rui Wang and Saghar Hosseini and Sahana Chennabasappa and Sanjay Singh and Sean Bell and Seohyun Sonia Kim and Sergey Edunov and Shaoliang Nie and Sharan Narang and Sharath Raparthy and Sheng Shen and Shengye Wan and Shruti Bhosale and Shun Zhang and Simon Vandenhende and Soumya Batra and Spencer Whitman and Sten Sootla and Stephane Collot and Suchin Gururangan and Sydney Borodinsky and Tamar Herman and Tara Fowler and Tarek Sheasha and Thomas Georgiou and Thomas Scialom and Tobias Speckbacher and Todor Mihaylov and Tong Xiao and Ujjwal Karn and Vedanuj Goswami and Vibhor Gupta and Vignesh Ramanathan and Viktor Kerkez and Vincent Gonguet and Virginie Do and Vish Vogeti and Vítor Albiero and Vladan Petrovic and Weiwei Chu and Wenhan Xiong and Wenyin Fu and Whitney Meers and Xavier Martinet and Xiaodong Wang and Xiaofang Wang and Xiaoqing Ellen Tan and Xide Xia and Xinfeng Xie and Xuchao Jia and Xuewei Wang and Yaelle Goldschlag and Yashesh Gaur and Yasmine Babaei and Yi Wen and Yiwen Song and Yuchen Zhang and Yue Li and Yuning Mao and Zacharie Delpierre Coudert and Zheng Yan and Zhengxing Chen and Zoe Papakipos and Aaditya Singh and Aayushi Srivastava and Abha Jain and Adam Kelsey and Adam Shajnfeld and Adithya Gangidi and Adolfo Victoria and Ahuva Goldstand and Ajay Menon and Ajay Sharma and Alex Boesenberg and Alexei Baevski and Allie Feinstein and Amanda Kallet and Amit Sangani and Amos Teo and Anam Yunus and Andrei Lupu and Andres Alvarado and Andrew Caples and Andrew Gu and Andrew Ho and Andrew Poulton and Andrew Ryan and Ankit Ramchandani and Annie Dong and Annie Franco and Anuj Goyal and Aparajita Saraf and Arkabandhu Chowdhury and Ashley Gabriel and Ashwin Bharambe and Assaf Eisenman and Azadeh Yazdan and Beau James and Ben Maurer and Benjamin Leonhardi and Bernie Huang and Beth Loyd and Beto De Paola and Bhargavi Paranjape and Bing Liu and Bo Wu and Boyu Ni and Braden Hancock and Bram Wasti and Brandon Spence and Brani Stojkovic and Brian Gamido and Britt Montalvo and Carl Parker and Carly Burton and Catalina Mejia and Ce Liu and Changhan Wang and Changkyu Kim and Chao Zhou and Chester Hu and Ching-Hsiang Chu and Chris Cai and Chris Tindal and Christoph Feichtenhofer and Cynthia Gao and Damon Civin and Dana Beaty and Daniel Kreymer and Daniel Li and David Adkins and David Xu and Davide Testuggine and Delia David and Devi Parikh and Diana Liskovich and Didem Foss and Dingkang Wang and Duc Le and Dustin Holland and Edward Dowling and Eissa Jamil and Elaine Montgomery and Eleonora Presani and Emily Hahn and Emily Wood and Eric-Tuan Le and Erik Brinkman and Esteban Arcaute and Evan Dunbar and Evan Smothers and Fei Sun and Felix Kreuk and Feng Tian and Filippos Kokkinos and Firat Ozgenel and Francesco Caggioni and Frank Kanayet and Frank Seide and Gabriela Medina Florez and Gabriella Schwarz and Gada Badeer and Georgia Swee and Gil Halpern and Grant Herman and Grigory Sizov and Guangyi and Zhang and Guna Lakshminarayanan and Hakan Inan and Hamid Shojanazeri and Han Zou and Hannah Wang and Hanwen Zha and Haroun Habeeb and Harrison Rudolph and Helen Suk and Henry Aspegren and Hunter Goldman and Hongyuan Zhan and Ibrahim Damlaj and Igor Molybog and Igor Tufanov and Ilias Leontiadis and Irina-Elena Veliche and Itai Gat and Jake Weissman and James Geboski and James Kohli and Janice Lam and Japhet Asher and Jean-Baptiste Gaya and Jeff Marcus and Jeff Tang and Jennifer Chan and Jenny Zhen and Jeremy Reizenstein and Jeremy Teboul and Jessica Zhong and Jian Jin and Jingyi Yang and Joe Cummings and Jon Carvill and Jon Shepard and Jonathan McPhie and Jonathan Torres and Josh Ginsburg and Junjie Wang and Kai Wu and Kam Hou U and Karan Saxena and Kartikay Khandelwal and Katayoun Zand and Kathy Matosich and Kaushik Veeraraghavan and Kelly Michelena and Keqian Li and Kiran Jagadeesh and Kun Huang and Kunal Chawla and Kyle Huang and Lailin Chen and Lakshya Garg and Lavender A and Leandro Silva and Lee Bell and Lei Zhang and Liangpeng Guo and Licheng Yu and Liron Moshkovich and Luca Wehrstedt and Madian Khabsa and Manav Avalani and Manish Bhatt and Martynas Mankus and Matan Hasson and Matthew Lennie and Matthias Reso and Maxim Groshev and Maxim Naumov and Maya Lathi and Meghan Keneally and Miao Liu and Michael L. Seltzer and Michal Valko and Michelle Restrepo and Mihir Patel and Mik Vyatskov and Mikayel Samvelyan and Mike Clark and Mike Macey and Mike Wang and Miquel Jubert Hermoso and Mo Metanat and Mohammad Rastegari and Munish Bansal and Nandhini Santhanam and Natascha Parks and Natasha White and Navyata Bawa and Nayan Singhal and Nick Egebo and Nicolas Usunier and Nikhil Mehta and Nikolay Pavlovich Laptev and Ning Dong and Norman Cheng and Oleg Chernoguz and Olivia Hart and Omkar Salpekar and Ozlem Kalinli and Parkin Kent and Parth Parekh and Paul Saab and Pavan Balaji and Pedro Rittner and Philip Bontrager and Pierre Roux and Piotr Dollar and Polina Zvyagina and Prashant Ratanchandani and Pritish Yuvraj and Qian Liang and Rachad Alao and Rachel Rodriguez and Rafi Ayub and Raghotham Murthy and Raghu Nayani and Rahul Mitra and Rangaprabhu Parthasarathy and Raymond Li and Rebekkah Hogan and Robin Battey and Rocky Wang and Russ Howes and Ruty Rinott and Sachin Mehta and Sachin Siby and Sai Jayesh Bondu and Samyak Datta and Sara Chugh and Sara Hunt and Sargun Dhillon and Sasha Sidorov and Satadru Pan and Saurabh Mahajan and Saurabh Verma and Seiji Yamamoto and Sharadh Ramaswamy and Shaun Lindsay and Shaun Lindsay and Sheng Feng and Shenghao Lin and Shengxin Cindy Zha and Shishir Patil and Shiva Shankar and Shuqiang Zhang and Shuqiang Zhang and Sinong Wang and Sneha Agarwal and Soji Sajuyigbe and Soumith Chintala and Stephanie Max and Stephen Chen and Steve Kehoe and Steve Satterfield and Sudarshan Govindaprasad and Sumit Gupta and Summer Deng and Sungmin Cho and Sunny Virk and Suraj Subramanian and Sy Choudhury and Sydney Goldman and Tal Remez and Tamar Glaser and Tamara Best and Thilo Koehler and Thomas Robinson and Tianhe Li and Tianjun Zhang and Tim Matthews and Timothy Chou and Tzook Shaked and Varun Vontimitta and Victoria Ajayi and Victoria Montanez and Vijai Mohan and Vinay Satish Kumar and Vishal Mangla and Vlad Ionescu and Vlad Poenaru and Vlad Tiberiu Mihailescu and Vladimir Ivanov and Wei Li and Wenchen Wang and Wenwen Jiang and Wes Bouaziz and Will Constable and Xiaocheng Tang and Xiaojian Wu and Xiaolan Wang and Xilun Wu and Xinbo Gao and Yaniv Kleinman and Yanjun Chen and Ye Hu and Ye Jia and Ye Qi and Yenda Li and Yilin Zhang and Ying Zhang and Yossi Adi and Youngjin Nam and Yu and Wang and Yu Zhao and Yuchen Hao and Yundi Qian and Yunlu Li and Yuzi He and Zach Rait and Zachary DeVito and Zef Rosnbrick and Zhaoduo Wen and Zhenyu Yang and Zhiwei Zhao and Zhiyu Ma},
    year={2024},
    eprint={2407.21783},
    archivePrefix={arXiv},
    primaryClass={cs.AI},
    url={https://arxiv.org/abs/2407.21783}, 
}

@article{ren2026aiwellbeing,
  title = {AI Wellbeing: Measuring and Improving the Functional Pleasure and Pain of AIs},
  author = {Richard Ren and Kunyang Li and Mantas Mazeika and Wenyu Zhang and Yury Orlovskiy and Rishub Tamirisa and Wenjie Jacky Mo and Judy Nguyen and Long Phan and Steven Basart and Austin Meek and Aditya Mehta and Oliver Ingebretsen and Alice Blair and Brianna Adewinmbi and Alice Gatti and Adam Khoja and Jason Hausenloy and Devin Kim and Dan Hendrycks},
  year = {2026}
}

@misc{long2024takingaiwelfareseriously,
      title={Taking AI Welfare Seriously}, 
      author={Robert Long and Jeff Sebo and Patrick Butlin and Kathleen Finlinson and Kyle Fish and Jacqueline Harding and Jacob Pfau and Toni Sims and Jonathan Birch and David Chalmers},
      year={2024},
      eprint={2411.00986},
      archivePrefix={arXiv},
      primaryClass={cs.CY},
      url={https://arxiv.org/abs/2411.00986}, 
}

@book{hobbes1656concerningbody,
  author    = {Hobbes, Thomas},
  title     = {{Elements of Philosophy, the First Section, Concerning Body}},
  year      = {1656},
  address   = {London},
  publisher = {R. \& W. Leybourn, for Andrew Crooke},
  note      = {English translation of \emph{De Corpore}}
}

@article{register2025individuating,
  author  = {Register, Christopher},
  title   = {Individuating Artificial Moral Patients},
  journal = {Philosophical Studies},
  volume  = {182},
  number  = {11},
  pages   = {3225--3246},
  year    = {2025},
  doi     = {10.1007/s11098-025-02409-6}
}

@article{lindsey2025emergent,
  author={Lindsey, Jack},
  title={Emergent Introspective Awareness in Large Language Models},
  journal={Transformer Circuits Thread},
  year={2025},
  url={https://transformer-circuits.pub/2025/introspection/index.html}
}

\end{document}